\documentclass[preprint,10pt]{elsarticle}

\usepackage[margin=1in]{geometry}
\pdfoutput=1 

\usepackage{amssymb,amsmath,amsthm,amsbsy}
\usepackage{physics}
\usepackage[pdftex,bookmarksnumbered]{hyperref}

\usepackage{amssymb,amsmath,amsthm,amsbsy}
\usepackage{systeme}
\usepackage{physics}
\usepackage{hyperref} 
\usepackage{graphicx}
\usepackage{epstopdf}
\usepackage{epsfig}
\usepackage{multicol}
\usepackage{subfigure}
\usepackage{enumitem}
\usepackage{multirow}
\usepackage{bm}
\usepackage{setspace}
\usepackage{color}
\usepackage[mathlines,running]{lineno}
\usepackage{tcolorbox}
\tcbuselibrary{skins,breakable}
\usepackage[hypcap=false]{caption}
\usepackage{todonotes}
\usepackage{fancyhdr}

\usepackage{tikz}
\newcommand*\circled[1]{\raisebox{.5pt}{\textcircled{\raisebox{-.9pt} {#1}}}}
\DeclareMathOperator*{\argmin}{arg\,min}

\begin{document}
\makeatletter
\def\ps@pprintTitle{%
  \let\@oddhead\@empty
  \let\@evenhead\@empty
  \let\@oddfoot\@empty
  \let\@evenfoot\@oddfoot
}
\makeatother


\begin{frontmatter}

\title{Multi-fidelity reduced-order surrogate modeling}




\author[add1,add1b]{Paolo Conti\corref{cor1}}
\ead{paolo.conti@polimi.it}
\cortext[cor1]{Corresponding author.}

\author[add2]{Mengwu Guo}
\ead{m.guo@utwente.nl}

\author[add3]{Andrea Manzoni}
\ead{andrea1.manzoni@polimi.it}

\author[add1]{Attilio Frangi}
\ead{attilio.frangi@polimi.it}

\author[add4,add1b]{Steven L. Brunton}
\ead{sbrunton@uw.edu}

\author[add5,add1b]{J. Nathan Kutz}
\ead{kutz@uw.edu}

\address[add1]{Department of Civil Engineering, Politecnico di Milano, Italy}
\address[add1b]{AI Institute in Dynamic Systems, University of Washington, Seattle, USA}
\address[add2]{Department of Applied Mathematics, University of Twente, Enschede, the Netherlands}
\address[add3]{MOX -- Department of Mathematics, Politecnico di Milano, Italy}
\address[add4]{Department of Mechanical Engineering, University of Washington, Seattle, USA}
\address[add5]{Department of Applied Mathematics and Electrical and Computer Engineering, University of Washington, Seattle, USA}

\begin{abstract}

\noindent 
High-fidelity numerical simulations of partial differential equations (PDEs) given a restricted computational budget can significantly limit the number of parameter configurations considered and/or time window evaluated for modeling a given system.   Multi-fidelity surrogate modeling aims to leverage less accurate, lower-fidelity models that are computationally inexpensive in order to enhance predictive accuracy when high-fidelity data are limited or scarce.  However, low-fidelity models, while often displaying important qualitative spatio-temporal features, fail to accurately capture the onset of instability and critical transients observed in the high-fidelity models, making them impractical as surrogate models.  To address this shortcoming, we present a new data-driven strategy that combines dimensionality reduction with multi-fidelity neural network surrogates.  The key idea is to generate a spatial basis by applying the classical proper orthogonal decomposition (POD) to high-fidelity solution snapshots, and approximate the dynamics of the reduced states --- time-parameter-dependent expansion coefficients of the POD basis -- using a multi-fidelity {\em long-short term memory} (LSTM) network.  By mapping low-fidelity reduced states to their high-fidelity counterpart, the proposed reduced-order surrogate model enables the efficient recovery of full solution fields over time and parameter variations in a non-intrusive manner.  The generality and robustness of this method is demonstrated by a collection of parametrized, time-dependent PDE problems where the low-fidelity model can be defined by coarser meshes and/or time stepping, as well as by misspecified physical features.  Importantly, the onset of instabilities and transients are well captured by this surrogate modeling technique.
\end{abstract}



\end{frontmatter}

\section{Introduction}
Scientific computing has revolutionized science and engineering by enabling advancements that have transformed almost every field of application.  While becoming an increasingly critical component of any real-world modeling, accurate and well-resolved simulations for multi-fidelity and multi-physics system also come at an elevated computational cost that can strain limited computational resources.  Indeed, it is often challenging to generate many high-fidelity (HF) simulations from large-scale models with limited availability of computing power, thus imposing restrictions on how comprehensive, either in parametric studies or length of time evolution, such numerical simulations can be.  In particular, computational costs can easily become prohibitive or intractable when parameterized, time-dependent systems of partial differential equations (PDEs) are solved with detailed full-order models (FOMs) in a multi-query context (i.e., at many instances of the input parameters characterizing the systems), such as in uncertainty quantification \cite{bui2008parametric, sudret2000stochastic},  optimal control \cite{negri2013reduced,sinigaglia2022fast}, shape optimization \cite{manzoni2012shape}, parameter estimation \cite{frangos2010surrogate,cui2015data}, and model calibration \cite{pateraYanoWD, haik2023real, rubio2021real}. In such cases, the construction of efficient surrogate models is of paramount importance in order to produce model proxies which can cheaply and accurately characterize the PDE system.  By mapping low-fidelity reduced states to their high-fidelity counterpart, we demonstrate a reduced-order surrogate model paradigm that enables the efficient recovery of full solution fields over time and parameter variations in a non-intrusive manner.

Reduced-order models (ROMs) have been developed to construct low-dimensional representations of high-dimensional systems for a significant reduction in computational costs with controlled accuracy \cite{QMN,benner2015survey,Antoulas2005,noack2011galerkin,Noack2011book,MDEIM,HRS}.  
Among the available strategies, \textit{intrusive} ROM techniques explicitly incorporate full-order governing equations at the reduced level and often yield reliable and physically meaningful solutions. However, the requirement for full-order simulators has limited the flexibility, generality, and industrial relevance of these approaches.  
On the other hand, \textit{non-intrusive} approaches learn reduced-order systems primarily from solution data, including from numerical or experimental data. Examples are dynamic mode decomposition \cite{schmid2010dynamic,brunton2022data}, reduced-order operator inference \cite{peherstorfer2016data,qian2020lift,ghattas2021learning, guo2022bayesian}, sparse identification of reduced latent dynamics \cite{brunton2016discovering,champion2019sindy, bakarji2022discovering, schaeffer2017learning, conti2023reduced,mars2022bayesian}, manifold learning using deep auto-encoders \cite{lee2020model, fresca2020comprehensive, fresca2021poddlrom, otto2019linearly}, data-driven approximation of time-integration schemes \cite{zhuang2021model}, and Gaussian processes for reduced representations \cite{guo2019data, botteghi2022deep}. These data-driven ROM techniques do not rely on direct operations on the full-order solvers, and are especially advantageous for applications with well-established, readily-executed legacy codes.

However, the applicability and reliability of these numerical methods can break down when the collection of HF data for model reduction is too computationally expensive, even in the offline stage of model training. In addition, it is increasingly common to encounter scenarios where a wealth of data sources are readily available, easily accessible, and/or cheaply computable, albeit not perfectly accurate. These low-fidelity (LF) data can be generated from coarse discretizations, linearization, simplified geometric or physical assumptions, or computationally efficient surrogate models. 
Despite limitations in accuracy, the LF data can represent a useful addition of information to the limited HF data used for model training. Thus, multi-fidelity (MF) methods aim to achieve an effective data fusion from various fidelity levels, and enable strong generalization performance of data-driven models in regions where HF data are scarce or even absent. A wide range of MF surrogate modeling techniques have been developed based on Gaussian processes \cite{kennedy2000predicting, alvarez2012kernels} and neural networks (NNs) \cite{meng2020composite, meng2020multi, liu2019multi, Motamed, guo2022multi}.  They have found recent applications in many areas of scientific computing, including uncertainty quantification, inference, and optimization \cite{peherstorfer2018survey, raissi2017inferring, pepper2021local, kast2020non, torzoni2023multi, geneva2020multi, ahmed2021multifidelity}.
Nevertheless, MF techniques often become impractical when approximating high-dimensional systems, thereby limiting their ability to directly approximate the full solution fields of PDEs. Fortunately, with the aid of dimensionality reduction, MF data fusion can be feasible for the representation of reduced states in a predominant low-dimensional latent space. 

\begin{figure}[t]
    \centering
    \includegraphics[width=1.\linewidth]{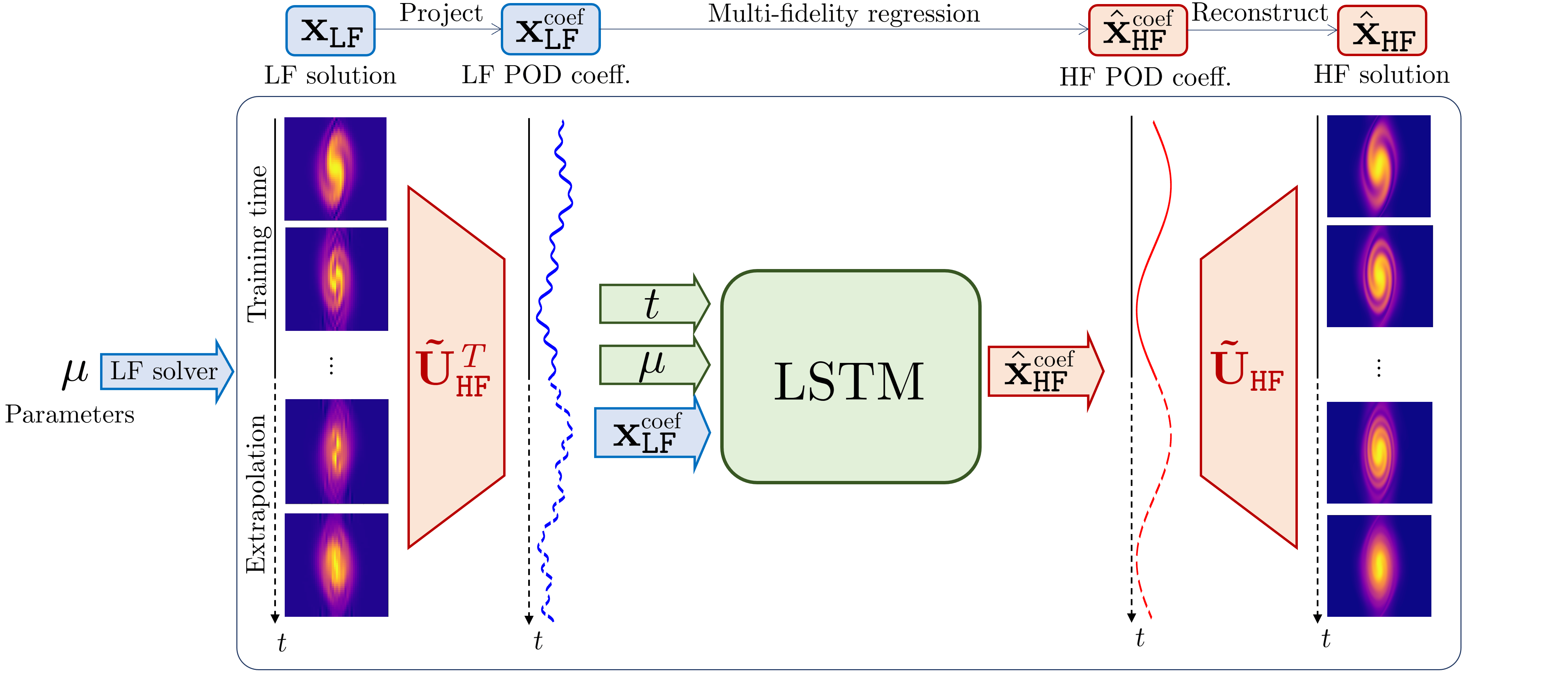}
    \caption{Using the proposed multi-fidelity reduced-order surrogate model to approximate time-parameter-dependent solution fields. Given a new parameter configuration $\mu$, low-fidelity solutions $\mathbf{x}_\texttt{LF}$ are evaluated over a desired time domain that can be much longer than the time window covered by training data. Low-fidelity POD coefficients ${\mathbf{x}}^\text{coef}_\texttt{LF}$ are computed via direct projection onto the reduced basis $\tilde{\mathbf{U}}_\texttt{HF}$, and their mapped to their high-fidelity counterpart $\hat{\mathbf{x}}^\text{coef}_\texttt{HF}$ through an LSTM neural network. Finally, the approximation of full solution fields is reconstructed as linear combinations of the reduced basis vectors.}
    \label{fig: test}
\end{figure}

To combine the computational flexibility of non-intrusive ROMs and data efficiency of MF modeling, we present an MF method of reduced-order surrogate modeling, abbreviated as MF-POD, which integrates MF regression with dimensionality reduction via the proper orthogonal decomposition (POD). The core idea is to approximate the solution manifold by a reduced subspace spanned by a small number of spatial bases using the POD, and then employ MF regression to represent their time-parameter-dependent expansion coefficients. The essence of MF regression here is inferring HF POD coefficients from their LF counterpart, so as to approach HF accuracy at the computational cost of LF evaluations. To achieve this, a time-parameter-dependent mapping from the LF to HF POD coefficients is constructed by means of long-short term memory (LSTM) NNs \cite{hochreiter1997long, gers2000learning}. LSTM models have been shown to be effective in time series analysis, especially for the detection of both long- and short-term temporal patterns and nonlinear correlations between datasets, with potential relevance in the construction of non-intrusive reduced order models \cite{ahmed2021multifidelity, vlachas2018data,vlachas2022multiscale,nakamura2021convolutional,maulik2021reduced}. A recent work \cite{conti2023multi} has also shown their success in MF surrogate modeling for simultaneous parametric generalization and temporal forecast. Thus, instead of relying on expensive HF full-order evaluations, MF-POD allows for efficient online approximation of solution fields over time and parameter variation by running fast LF simulations and then mapping their POD coefficients to the HF level. 
In this way, parameter regions with sparse (or even no) HF data coverage can be conveniently explored.  From the time evolution of the LF model, the long-term HF forecast is enabled through LSTM models. A schematic representation of the method is represented in Fig.~\ref{fig: test}.

The major advantage of the proposed method lies in a guaranteed light-weight offline stage. The computational cost of MF data generation is reduced, as the need for HF samples (for both the POD and LSTM training) is limited and the computation of LF samples (for LSTM training) is extremely efficient. The POD reduction also ensures that the LSTM time series is modeled in a relatively low dimension. 
While leveraging the advantages of being non-intrusive, MF-POD overcomes the potential lack of physical consistency by incorporating physically meaningful LF data, hence enabling interpretability and reliability in generalization.
We test MF-POD's performance in parametric generalization and temporal forecasting on a diverse set of PDE benchmark problems, including spiral wave propagation in a parametrized reaction-diffusion system, vorticity approximation of advection-diffusion in shallow water, and velocity and pressure approximation of fluid flow past a cylinder in a channel. Compared to detailed HF simulations, LF data are generated by lower-quality discretizations over space and time, and/or with corrupted values of critical physical features.


This paper is structured as follows. In section \ref{sect: train}, we introduce the \textit{offline} training process of the proposed multi-fidelity reduced-order surrogate model, which consists of the POD reduced basis construction and the LF-to-HF mapping with LSTM network models, both based on bi-fidelity data at limited time-parameter locations; thereafter, using this trained model, we present the \textit{online} procedure to infer HF solutions from LF evaluations over a wider range of parametric configurations and forward in time. Results for the aforementioned numerical tests are reported and discussed in sections  \ref{sect: results1} to \ref{sect: results3}, and conclusions are finally drawn in section \ref{sect: conclusions}. The source code of the proposed method is made available in the public repository \texttt{MultiFidelity\_POD} \cite{MFPOD_repo}.

\section{MF-POD: offline/online framework}
\label{sect: train}
In this section, we introduce an algorithmic method for MF reduced-order surrogate modeling --- MF-POD, which is decoupled into the stages of offline training and online testing.

\subsection{Offline training}
\label{sec: off train}

\begin{figure}[t!]
    \centering
    \includegraphics[width=1.\linewidth]{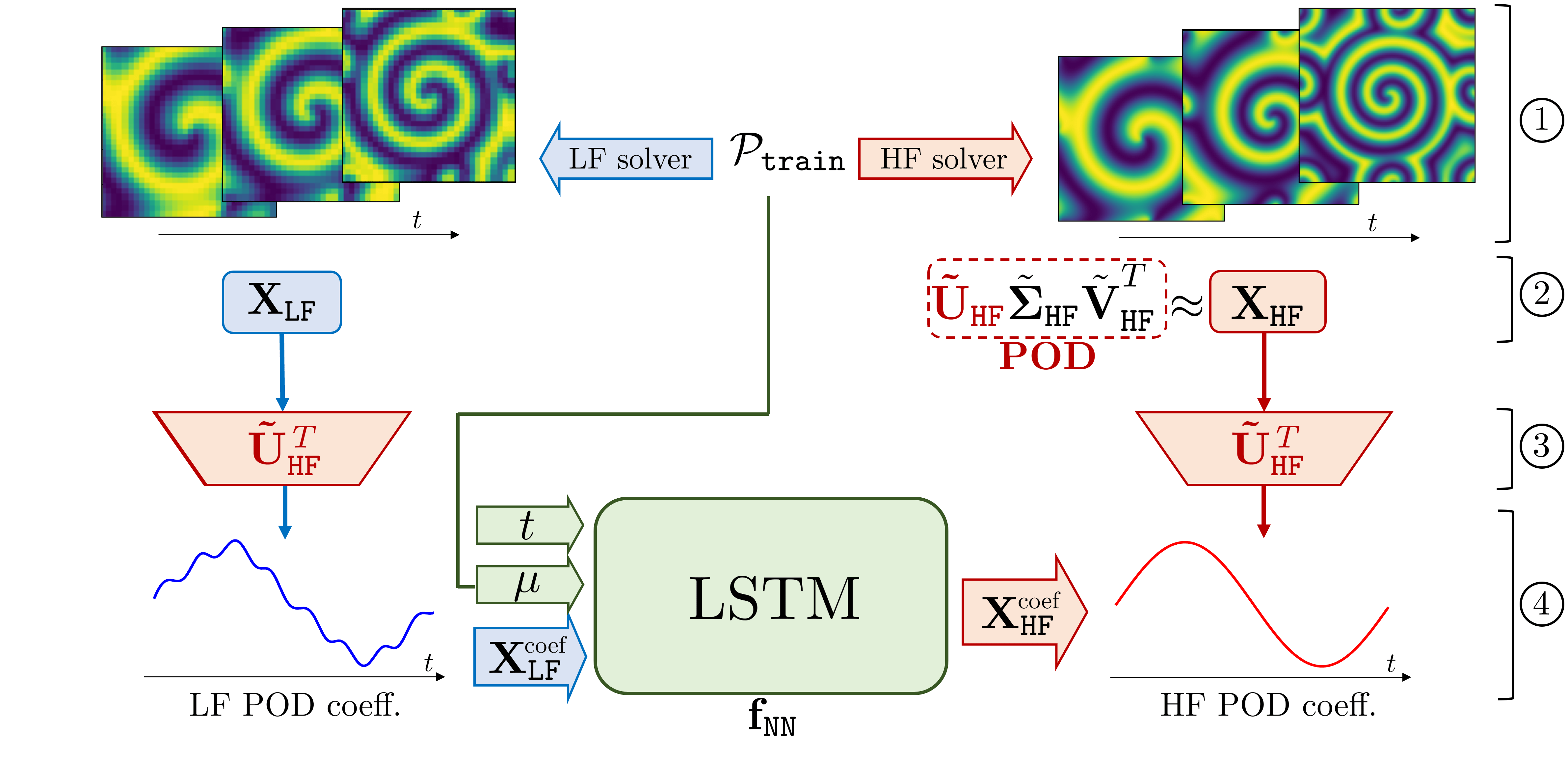}
    \caption{Offline training process of the proposed MF-POD strategy. The steps are \circled{1} generating multi-fidelity training datasets, \circled{2} dimensionality reduction via POD, \circled{3} computing POD coefficients via direct projection, and \circled{4} training LSTM surrogate model.}
    \label{fig:train}
\end{figure}

A schematic presentation of the offline training is illustrated in Fig.~\ref{fig:train}, while its algorithmic implementation is outlined in what follows.

\medskip
\noindent $\bullet$~\textbf{Step 1: Generating MF training datasets}\\
HF and LF solution data for training are computed over a limited set of parametric configurations $\mathcal{P}_\text{train}$ by running the respective solvers. HF (resp. LF) snapshots are stacked in a matrix $\textbf{X}_\texttt{HF}\in \mathbb{R}^{N_\text{dof}^\texttt{HF}\times N_\mu N_t^\texttt{HF}}$ (resp. $\textbf{X}_\texttt{LF}\in \mathbb{R}^{N_\text{dof}^\texttt{LF}\times N_\mu N_t^\texttt{LF}}$), where $N_\mu = |\mathcal{P}_\text{train}|$, and $N_\text{dof}^\texttt{HF}$ (resp. $N_\text{dof}^\texttt{LF}$) is the number of spatial degrees of freedom and $N_t^\texttt{HF}$ (resp. $N_t^\texttt{LF}$) the number of time instances for the HF (resp. LF) level. Note that solutions at different fidelity levels are evaluated at the same parameter values, but not necessarily at the same spatio-temporal locations, because the latter depends on the spatial mesh and time stepping of choice. This step demands the highest computational cost in the offline stage, as it requires querying the HF full-order model. Therefore, we consider a small number of parameter instances $N_\mu$ for the training data. 

\medskip
\noindent $\bullet$~\textbf{Step 2: Dimensionality reduction via POD}\\ 
A set of $N_\texttt{POD}$ reduced basis vectors, collected in $\tilde{\textbf{U}}_\texttt{HF}\in\mathbb{R}^{N_\text{dof}^\texttt{HF} \times N_\texttt{POD}}$, is extracted from the HF snapshot matrix $\textbf{X}_\texttt{HF}$ via the POD. Note that $\tilde{\textbf{U}}_\texttt{HF}^T\tilde{\textbf{U}}_\texttt{HF}=\mathbf{I}_{N_\texttt{POD}}$.

\begin{tcolorbox}[enhanced, breakable]
	\textbf{Proper Orthogonal Decomposition (POD)}: The POD takes advantage of the singular-value decomposition (SVD) to linearly extract principal components from high-dimensional data, and hence provide low-dimensional orthonormal basis that can be computed as follows:

    \begin{enumerate}
        \item Compute the SVD of $\textbf{X}_\texttt{HF}$, such that
        \begin{equation}
            \textbf{X}_\texttt{HF} = \textbf{U}_\texttt{HF}\mathbf{\Sigma}_\texttt{HF}\textbf{V}^T_\texttt{HF}, ~~~ \text{with}~~~ 
            \mathbf{\Sigma}_\texttt{HF} = \text{diag}(\sigma_1, \ldots, \sigma_{ N_\mu N_t^\texttt{HF}})~~~ \text{and}~~~ 
            \textbf{U}_\texttt{HF}^T\textbf{U}_\texttt{HF}=\mathbf{I},~~ \textbf{V}_\texttt{HF}^T\textbf{V}_\texttt{HF}=\mathbf{I}\,.
        \end{equation}
        Here $\sigma_1 \geq \sigma_2 \geq \ldots \geq \sigma_{N_\mu N_t^\texttt{HF}} \geq 0$.
    \item Define $N_\texttt{POD}$ as the minimum integer that satisfies
    \begin{equation}
        {\frac{\sum_{i=1}^{N_\texttt{POD}} \sigma_i^2}{\sum_{i=1}^{N_\mu N_t^\texttt{HF}} \sigma_i^2}}\geq 1-\epsilon^2_\text{POD},
    \end{equation}
    where $\epsilon_\texttt{POD}>0$ is a given tolerance that determines how much of the variance of the signal should be captured.
    As an alternative to providing a fixed tolerance $\epsilon_\texttt{POD}$, one can select the dimension $N_\texttt{POD}$ of the reduced basis according to the singular-value decay.
    \item Form the POD basis $\tilde{\mathbf{U}}_\texttt{HF}$ as the first $N_\texttt{POD}$ columns of $\mathbf{U}_\texttt{HF}$, i.e., $\tilde{\mathbf{U}}_\texttt{HF} = \left(\mathbf{U}_\texttt{HF}\right)_{:,1:N_\texttt{POD}}$.
    \end{enumerate}

    The POD truncation provides a low-rank approximation of $\textbf{X}_\texttt{HF}$, i.e., $\textbf{X}_\texttt{HF} = \textbf{U}_\texttt{HF}\mathbf{\Sigma}_\texttt{HF}\textbf{V}^T_\texttt{HF} \approx \tilde{\textbf{U}}_\texttt{HF}\tilde{\mathbf{\Sigma}}_\texttt{HF}\tilde{\textbf{V}}^T_\texttt{HF}$, where $\tilde{\textbf{V}}_\texttt{HF}$ contains the first $N_\texttt{POD}$ columns of $\textbf{V}_\texttt{HF}$ and $\tilde{\mathbf{\Sigma}}_\texttt{HF}$ contains the first $N_\texttt{POD} \times N_\texttt{POD}$ block of $\mathbf{\Sigma}_\texttt{HF}$.
\end{tcolorbox}

\medskip
\noindent $\bullet$~\textbf{Step 3: Computing POD coefficients via direct projection}\\
\noindent The time-parameter-dependent combination coefficients of the POD basis for both fidelity levels are obtained by projecting $\textbf{X}_\texttt{HF}$ and $\textbf{X}_\texttt{LF}$ onto the basis $\tilde{\textbf{U}}_\texttt{HF}$, respectively:
\begin{equation}
  {\textbf{X}}^\text{coef}_\texttt{HF} = \tilde{\textbf{U}}_\texttt{HF}^T \textbf{X}_\texttt{HF}, \qquad {\textbf{X}}^\text{coef}_\texttt{LF} = \tilde{\textbf{U}}_\texttt{HF}^T \mathcal{L}(\textbf{X}_\texttt{LF}),  
\end{equation}
where $\textbf{X}^\text{coef}_\texttt{HF}, \textbf{X}^\text{coef}_\texttt{LF} \in \mathbb{R}^{N_\texttt{POD}\times N_\mu N_t^\texttt{HF}}$ collect respectively the POD coefficients of the HF and LF data. Here, $\mathcal{L}$ represents an operator which lifts the LF solution vectors to the HF spatio-temporal resolution via interpolation. In particular, spatial interpolation is required when the LF model is on a different (coarser) mesh than the HF one, thus the LF solution vectors should be transformed to match with the HF resolution and hence enable a projection onto the POD modes $\tilde{\textbf{U}}_\texttt{HF}$; temporal interpolation is necessary when LF and HF data are computed with different time discretizations, because compatible sequential data between the two fidelity levels are required by the NN model in the next step. 
In this work, we consider either linear interpolation (i.e.,  $\mathcal{L}(\textbf{X}_\texttt{LF}) = \textbf{P}\textbf{X}_\texttt{LF}\textbf{Q}^T$ with $\textbf{P}\in\mathbb{R}^{N_\text{dof}^{\texttt{HF}} \times N_\text{dof}^{\texttt{LF}} }$ and $\textbf{Q}\in\mathbb{R}^{N_\mu N_t^{\texttt{HF}}\times N_\mu N_t^{\texttt{LF}} }$ being the spatial and temporal interpolating matrices, respectively) or nearest-neighbor interpolation, both of which are computationally inexpensive. An alternative option for matching LF and HF solution vectors is manifold alignment \cite{wang2009general, perron2020development}.

\medskip
\newpage
\noindent $\bullet$~\textbf{Step 4: Training LSTM surrogate model}\\
The goal of this step is to learn the mapping $ \mathbf{f}$ from the LF POD coefficients ${\mathbf{x}}^\text{coef}_\texttt{LF} \in \mathbb{R}^{N_\texttt{POD}}$ at time instance $t$ and parameter $\mu$ to their HF correspondence ${\mathbf{x}}_\texttt{HF}^\text{coef} \in \mathbb{R}^{N_\texttt{POD}}$, written as
\begin{equation}
    \left(t,\mu,{\mathbf{x}}_\texttt{LF}^\text{coef}(t,\mu)\right)\mapsto \mathbf{f}(t,\mathbf{\mu},{\mathbf{x}}^\text{coef}_\texttt{LF}(t,\mu))={\mathbf{x}}_\texttt{HF}^\text{coef}(t,\mu).
\end{equation}
To approximate the mapping $\mathbf{f}$, we make use of a long short-term memory (LSTM) NN model denoted by $\mathbf{f}_\texttt{NN}(\cdot) = \mathbf{f}_\texttt{NN}(\cdot; \mathbf{\Theta}_\texttt{NN})$, where $\mathbf{\Theta}_\texttt{NN}$ are the network parameters. 
This NN model $\mathbf{f}_\texttt{NN}$ is trained on the input-output pairs of POD coefficients $\{({\vb{x}}^\text{coef}_\texttt{LF}(t_i,\mu_i),{\vb{x}}^\text{coef}_\texttt{HF}(t_i,\mu_i) )\}_{i=1}^{N_\mu N_t^\texttt{HF}}\in \mathbb{R}^{N_\texttt{POD}}$ (the columns of the matrices ${\textbf{X}}^\text{coef}_\texttt{LF}$ and ${\textbf{X}}^\text{coef}_\texttt{HF}$, respectively, for each time-parameter instance $(t_i,\mu_i)$). Using the Adam \cite{kingma2014adam} algorithm, the LF to HF mapping $\mathbf{f}_\texttt{NN}$ is determined by minimizing the mean squared error loss function as follows:
\begin{equation}
\label{eq: loss}
\mathbf{\Theta}_\texttt{NN}= \argmin_{\mathbf{\Theta}_\texttt{NN}}\; \frac{1}{N_\mu N_t^\texttt{HF}}\sum_{i=1}^{N_\mu N_t^\texttt{HF}}
\norm{{\mathbf{x}}^\text{coef}_\texttt{HF}(t_i,\mu_i) - \mathbf{f}_\texttt{NN}(t_i,\mu_i,{\mathbf{x}}^\text{coef}_\texttt{LF}(t_i,\mu_i); \mathbf{\Theta}_\texttt{NN}) }^2_{2}\,. 
\end{equation}    

\begin{tcolorbox}[enhanced, breakable]
\textbf{Long Short-Term Memory (LSTM) NNs for MF regression:} 
In this work, an LSTM network model is used for the time evolution of HF POD coefficients ${\mathbf{x}}^\text{coef}_\texttt{HF}$ by providing the corresponding sequence of LF coefficients ${\mathbf{x}}^\text{coef}_\texttt{LF}$ in the inputs together with the parameters $\mu$ and time $t$. As shown in Fig. \ref{fig:LSTM}, an LSTM unit has two latent recurrent states $\mathbf{c}$ and $\mathbf{h}$. At each time-step $n$, $(\vb{c}_n, \vb{h}_n)$ are recurrently updated from the combination of their last-step states $(\vb{c}_{n-1}, \vb{h}_{n-1})$ and the input values $\vb{x}_n = [t_n, \mu, {\mathbf{x}}^\text{coef}_\texttt{LF}(t_n,\mu)]$ through a three-fold \textit{gate mechanism} with a \textit{forget} gate $\vb{G}_f$, a \textit{update} gate $\vb{G}_u$, and a \textit{output} gate $\vb{G}_o$. The current state ${\mathbf{h}}_n \equiv \mathbf{f}_\texttt{NN}(t,\mu,{\mathbf{x}}^\text{coef}_\texttt{LF}(t_n,\mu))$ is produced as the unit's output $\vb{y}_n$ to approximate ${\mathbf{x}}^\text{coef}_\texttt{HF}(t_n,\mu)$ (refer to \eqref{eq: loss}).  
The gate mechanism is the core of an LSTM unit, designed to allow a refined memory management by weighting the contribution of past and present information, and to overcome major drawbacks of recurrent networks such as exploding and vanishing gradients. The recurrent states are updated as follows:
\begin{equation*} 
\label{eq: cnhn}
    \vb{c}_n = \vb{G}_f \circ \vb{c}_{n-1} + \vb{G}_u \circ \Tilde{\vb{c}}_n \,,\quad 
    \vb{h}_n = \vb{G}_o \circ \tanh{\vb{c}_n}\,\equiv \vb{y}_n\,, 
\end{equation*}
in which the operation $\circ$ denotes an element-wise product, $\Tilde{\vb{c}}_n = \tanh{(\vb{W}_c [\vb{h}_{n-1}, \vb{x}_{n}] + \vb{b}_c)}$ represents the new candidate  state to replace ${\vb{c}}_{n-1}$, and the gates $\{\vb{G}_i\}_{i \in \{f,g,u\}}$ are defined as $\vb{G}_i = \sigma(\vb{W}_i [\vb{h}_{n-1}, \vb{x}_{n}] + \vb{b}_i)$.
Here, $\{\vb{W}_c,\vb{W}_f,\vb{W}_u,\vb{W}_o\}$ and $\{\vb{b}_c, \vb{b}_f,\vb{b}_u,\vb{b}_o\}$ are the trainable parameters -- weights and biases -- of the LSTM unit to be included in $\mathbf{\Theta}_\texttt{NN}$, and $\sigma$ denotes the sigmoid activation function.

\includegraphics[width=.85\linewidth]{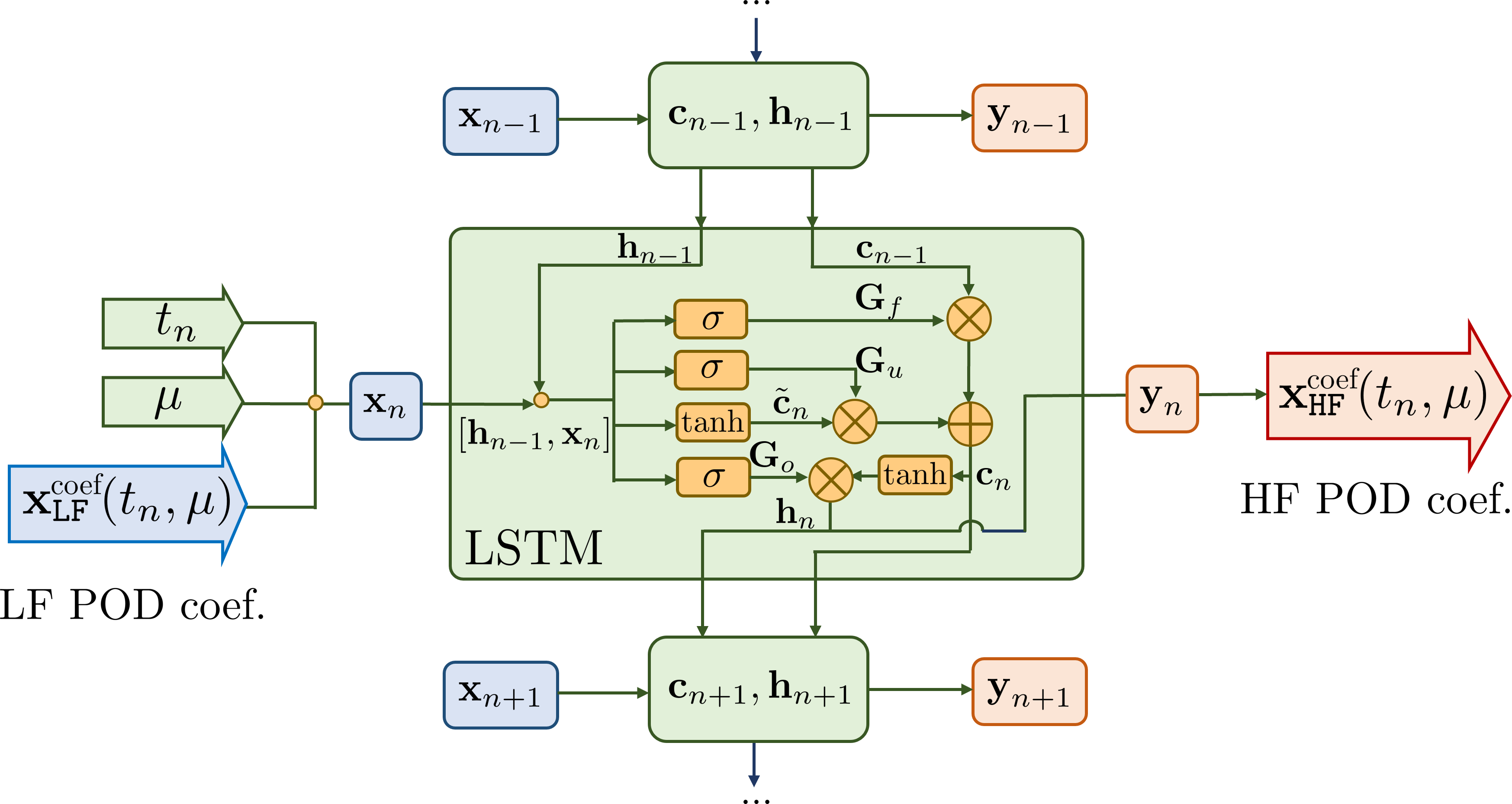}
\centering
\captionof{figure}{The visualization of an LSTM unit with its input-output setting \cite{olah2015understanding, conti2023multi}.}
\label{fig:LSTM}
\end{tcolorbox}

\smallskip
Since LSTM units are recurrent NNs that deal with sequential data, in practice, training data are grouped in batch subsequences with shape $n_\text{batch} \times K \times N_\texttt{POD}$, in which $n_\text{batch}$ is the batch size and $K$ is the length of batch subsequences. To determine the NN model's hyperparameters, including the network architecture (i.e., the numbers of layers and  nodes in each layer) and the parameters associated to model training (e.g., the optimizer learning rate), we use a Bayesian optimization technique \cite{bergstra2011algorithms} implemented by the Python package Hyperopt \cite{bergstra2022hyperopt}.

\subsection{Online testing}
\label{sect: online_test}

Once the MF-POD model is trained offline, it can be used online to efficiently evaluate PDE solutions for unseen parameter locations and longer time horizons, all at a very limited computational cost. As illustrated in Fig.~\ref{fig: test}, for new parameter configurations of interest, we compute the LF solution $\mathbf{x}_\texttt{LF}$ and project it onto the POD basis to obtain the corresponding reduced states given by the projection coefficients ${\mathbf{x}}^\text{coef}_\texttt{LF}$. These coefficients are passed as inputs to the MF LSTM model and hence mapped to the sequences of POD coefficients towards the HF level, obtaining $\hat{\mathbf{x}}^\text{coef}_\texttt{HF}$, from which the high-resolution fields $\hat{\mathbf{x}}_\texttt{HF}$ can be reconstructed. Thus, the proposed reduced-order surrogate model allows to provide an approximation of HF simulations at a cheap online price of LF run-times and, moreover, to infer the future states of HF solutions from the LF solutions evaluated forward in time. This enables reliable long-term forecast without querying the expensive full-order model at all.

\subsection{Metrics for performance evaluation}
\label{sect: metrics}
In the following sections we will assess the performance of MF-POD in several numerical examples. To highlight the advantages of MF-POD over HF solvers in saving computational costs, as well as the gain in accuracy in comparison to LF solvers, we report here the metrics for performance evaluation in terms of computational time and errors.
\setlength{\leftmargini}{10pt}
\begin{itemize}
    \item Computational time is recorded for the complete time evolution of interest, averaged over the testing parameter instances. For LF and HF solutions, this means the run time of respective solvers, while for the MF model, it refers to the evaluations in the \textit{online testing} procedure. Percentages are computed with respect to the HF time.
    \item Relative error with respect to the HF reference solution $\mathbf{x}_\texttt{HF}$ is evaluated for both the MF solution $\hat{\mathbf{x}}_\texttt{HF}$ and the \emph{lifted LF input}  $\mathbf{x}_\texttt{LF}$, as an average over the test set:
    \begin{equation}
    \begin{aligned}
         \text{err}^{\%}_\text{MF-POD}= \frac{100\%}{N_\text{test} }\sum_{i=1}^{N_\text{test} } \frac{
        \norm{{\mathbf{x}}_\texttt{HF}(t_i,\mu_i) - \hat{\mathbf{x}}_\texttt{HF}(t_i,\mu_i) }_{2}}{\norm{{\mathbf{x}}_\texttt{HF}(t_i,\mu_i)}_{2}}
        \,, \;
        \text{err}^{\%}_\text{LF}  = \frac{100\%}{N_\text{test} }\sum_{i=1}^{N_\text{test} } \frac{
        \norm{{\mathbf{x}}_\texttt{HF}(t_i,\mu_i) - {\mathbf{x}}_\texttt{LF}(t_i,\mu_i) }_{2}}{\norm{{\mathbf{x}}_\texttt{HF}(t_i,\mu_i)}_{2}}\, ,
    \end{aligned}    
    \end{equation}
    where $N_\text{test}$ is the number of time-parameter combinations in the test set.
\end{itemize}

\section{Numerical example I: Reaction-diffusion problem}
\label{sect: results1}

We consider a lambda-omega reaction-diffusion system governed by the following equations
\begin{equation}
    \begin{aligned}
&\dot{u}=\left(1-\left(u^{2}+v^{2}\right)\right) u+\mu\left(u^{2}+v^{2}\right) v+d\left(u_{x x}+u_{y y}\right),\\ 
&\dot{v}=-\mu\left(u^{2}+v^{2}\right) u+\left(1-\left(u^{2}+v^{2}\right)\right) v+d\left(v_{x x}+v_{y y}\right) , 
\end{aligned}
\label{eq: RD_eqs}
\end{equation}
defined over a spatial domain $(x,y)\in [-L,L]^2$ for $L=20$ and a time span $t\in [0,T]$ for $T=80$, where $\mu$ and $d$ are parameters that respectively regulate the reaction and diffusion behaviors of the system. 
We prescribe periodic boundary conditions, and the initial condition is defined as
\begin{equation*}
  \begin{aligned}
     u(x,y,0) = v(x,y,0) = \tanh{\left(\sqrt{x^2 + y^2} \cos{\left((x+iy)-\sqrt{x^2 + y^2}\right)}\right)}\, . 
  \end{aligned}
\end{equation*}
The solution $[u,v]^T(x,y,t)$ to problem \eqref{eq: RD_eqs} represents two oscillating modes which generate spiral waves, representing an attracting limit cycle in the state space. 

Our goal is to approximate the solution components $u$ and $v$ as functions of the varying reaction parameter $\mu\in \mathcal{P} =[0.5, 1.5]$ with a fixed diffusion coefficient $d = 0.05$. 
We employ the MF-POD method to efficiently evaluate high-resolution solutions over the whole time span with certain parametric variation, with the aid of cheaply obtained, low-resolution LF approximations. To reduce offline computational costs, we train the MF-POD model with a small amount of expensive HF solution data computed at a limited set of parameter locations over a shorter time horizon $T_\text{train}= 40 <T$, while leveraging LF solutions generated on a coarse spatial mesh with a corrupted value of diffusion coefficient. 

\subsection{Multi-fidelity setting}

\begin{figure}[t]
    \centering
    \includegraphics[width=0.85\linewidth]{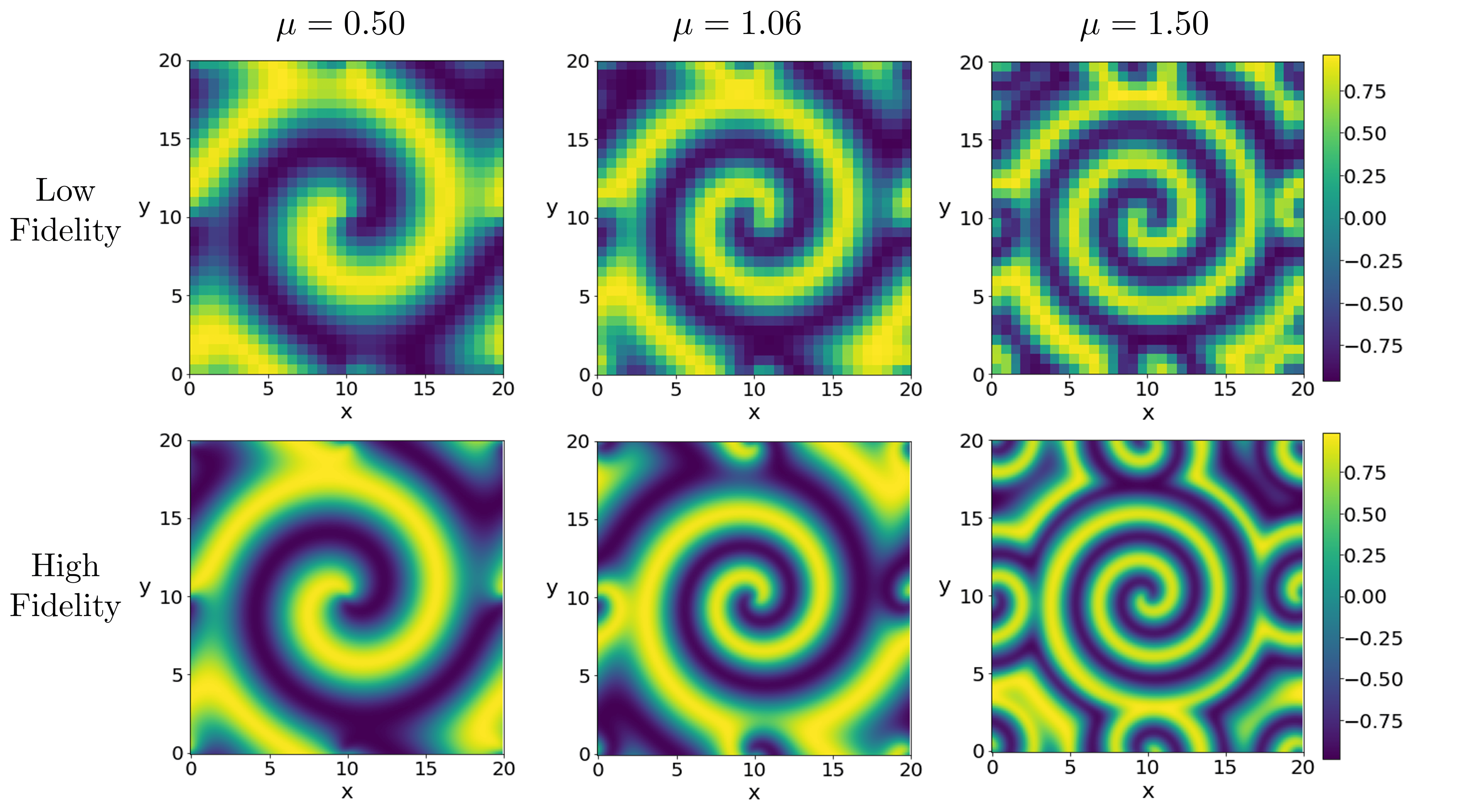}
    \caption{LF (above) and HF (below) solution snapshots of $u$ at three parameter locations and $t=T_\text{train} = 40$ in example (I). Besides the difference in spatial resolution, the corrupted diffusion coefficient in the LF model leads to different shapes of spiral waves than those in the HF solutions.}
    \label{fig:data_RD}
\end{figure}

Both HF and LF solution datasets are constructed by solving the PDEs \eqref{eq: RD_eqs} using the Fourier spectral method \cite{trefethen2000spectral} with time step $\Delta t = 0.05$. The two fidelity levels are defined as follows, and the difference between LF and HF solutions is shown in Fig. \ref{fig:data_RD}.

\begin{itemize}
    \item[-] LF solution data are generated on a coarse equispaced spatial grid with $n_\texttt{LF} = 32$ points in each direction, while a fine grid with $n_\texttt{HF} = 100$ is adopted for the HF data. 
    \item[-] LF solutions are evaluated at a corrupted diffusion coefficient $d_\texttt{LF} = 0.1$, instead of $d_\texttt{HF} = d = 0.05$. This represents a bias in the LF modeling in terms of the physical property of viscosity.
    \item[-] HF data are only available over a limited time window $[0,T_\text{train}]$ with $T_\text{train} = 40<T=80$.
    We hence aim to extrapolate for a same-length time window beyond that covered by the HF training data.
    \item[-] Training data on both fidelity levels are computed for a small number of parameter instances $\mu \in \mathcal{P} =[0.5, 1.5]$. In particular, $N_\mu = 10$ $\mu$-values are selected over an equispaced grid of $\mathcal{P}$. 
\end{itemize}

To apply the MF-POD method, we perform the POD reduction on the HF snapshots and retain the first $N_\texttt{POD} = 9$ modes. LF data are lifted to the HF spatial dimensionality via nearest-neighbor interpolation. For both fidelity levels, POD coefficients are computed by projecting the data onto the reduced basis and then fed to the LSTM neural network for training. Once the training phase is concluded, we run the LF solvers to efficiently evolve LF solutions over the whole time window $[0,T]$, and test the MF-POD accuracy in estimating the HF solutions over an unseen set of $N_\mu^\texttt{test} = 25$ equispaced parameter locations over $\mathcal{P}$.

\subsection{Results}

\begin{figure}[t]
    \centering
    \includegraphics[width=1.\linewidth]{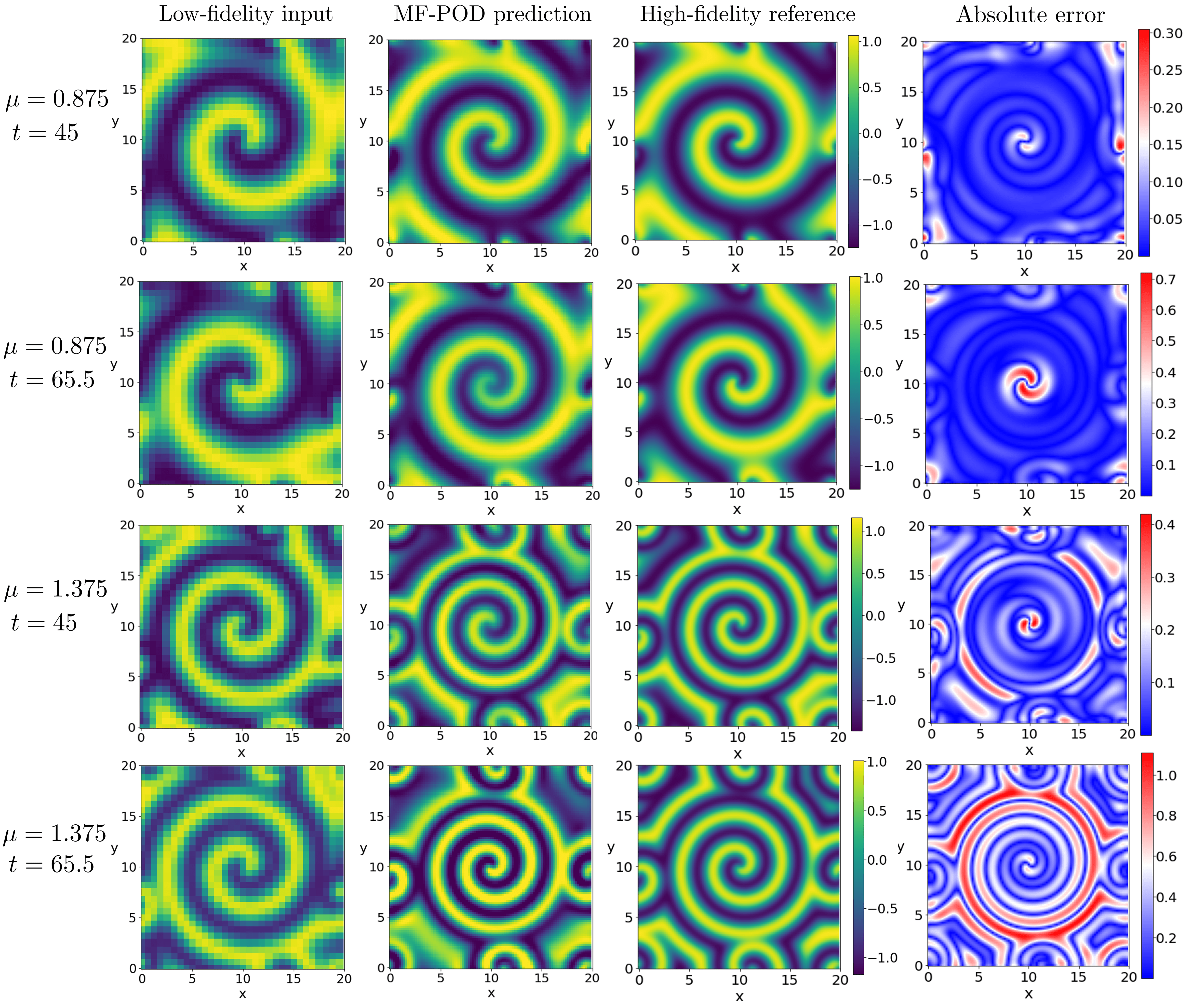}
    \caption{Comparison of solution fields in example (I) among the approximation by MF-POD, the corresponding LF input, and the HF ground truth (used as reference). The snapshots refer to two extrapolated time instances $t \in \{45, 65.5\}$ (being $T_\text{train} = 40$ the end of HF training coverage) for two testing values of reaction parameters $\mu \in \{0.875, 1.375\}$ that are unseen during the training. Absolute error shows the discrepancy between the MF solution and the HF reference.}
    \label{fig:physical_rec_RD}
\end{figure}

For a few extrapolated time instances at unseen testing values of the reaction coefficient $\mu$, we show in Fig.~\ref{fig:physical_rec_RD} the reconstruction of entire solution field predicted by MF-POD, compared with both HF reference and LF input. The proposed MF strategy is able to recover high resolution and correct the prediction of system behavior from inaccurate LF solutions. Table \ref{tab: tab_ex1} provides a quantitative comparison of computational time and relative errors among LF, HF, and MF solutions, highlighting the good performance of MF-POD.

\begin{figure}[ht]
    \centering
    \includegraphics[width=1.\linewidth]{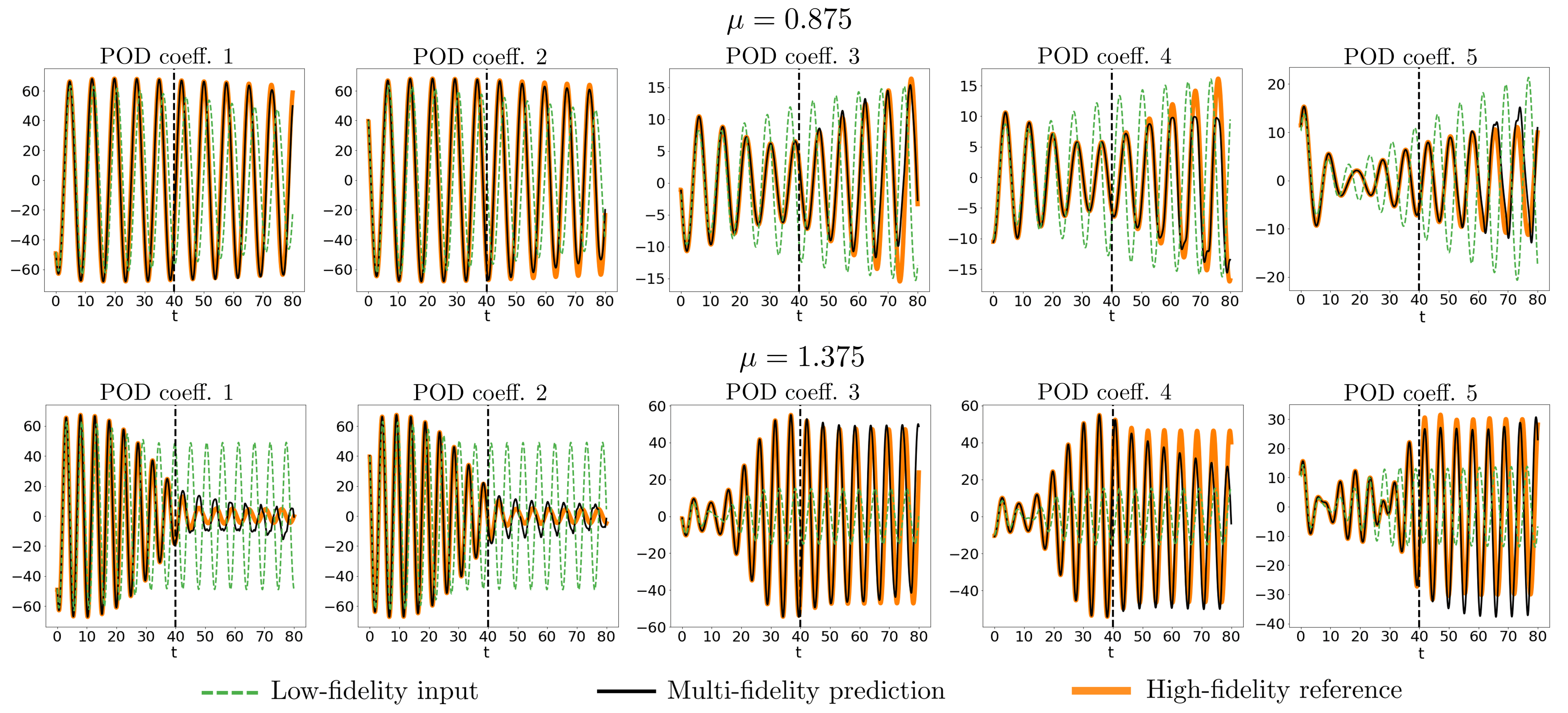}
    \caption{Comparison of the time evolution of POD coefficients in example (I) among MF solution, LF input, and HF reference at testing parameter instances $\mu = 0.875$ (above) and  $1.375$ (below). The dashed line at $T_\text{train} = 40$ indicates the end time of HF data coverage in training, i.e., no HF information is available over $t\in[T_\text{train},T] = [40,80]$.}
    \label{fig:temporal_coeff_RD}
\end{figure}

The combination of POD reduction and MF regression with LSTM proves to be effective in this example. The POD at the HF level extracts a global basis that represents predominant spatial patterns of the spiral wave propagation, and the LSTM network model approximates the corresponding expansion coefficients that describe how these spatial modes evolve over time. The expressive power of LSTM neural networks is crucial not only for the approximation of POD coefficients' strongly nonlinear dependency on time and parameters, but also for the correlation modeling between the LF and HF levels, especially considering that such an unknown correlation is parameter-dependent and potentially highly complex.
In Fig.~\ref{fig:temporal_coeff_RD}, we depict the evolution of several POD coefficients predicted by the LSTM network.
We notice that MF-POD provides very accurate estimation of the time-dependent POD coefficients at unseen parameter locations, not only in the training time interval $[0,40]$, but also for the future states over $[40,80]$, as the MF-predicted coefficients match very well with the HF reference and present a significant improvement compared to the LF level. These results over twice the length of the training time span indicate the capability of MF-POD in long-term forecast.

\begin{table}[b]
\caption{Comparison of computational time and accuracy in example (I). See section \ref{sect: metrics} for the definition of metrics.}
\centering
\begin{tabular}{c|c|c|c}
\hline
& Low-fidelity input & MF-POD predicted  & High-fidelity reference \\
\hline\hline
Computational time &    0.58s (2.70\%)          &      1.89s (8.79\%)        &     21.49s (100\%)       \\ 
\hline
Relative error           & 111\%      &      16\%          &      -         \\ \hline
\end{tabular}
    \label{tab: tab_ex1}
\end{table}

\subsection{The need of LSTM networks}
\begin{figure}
    \centering
    \includegraphics[width=1.\linewidth]{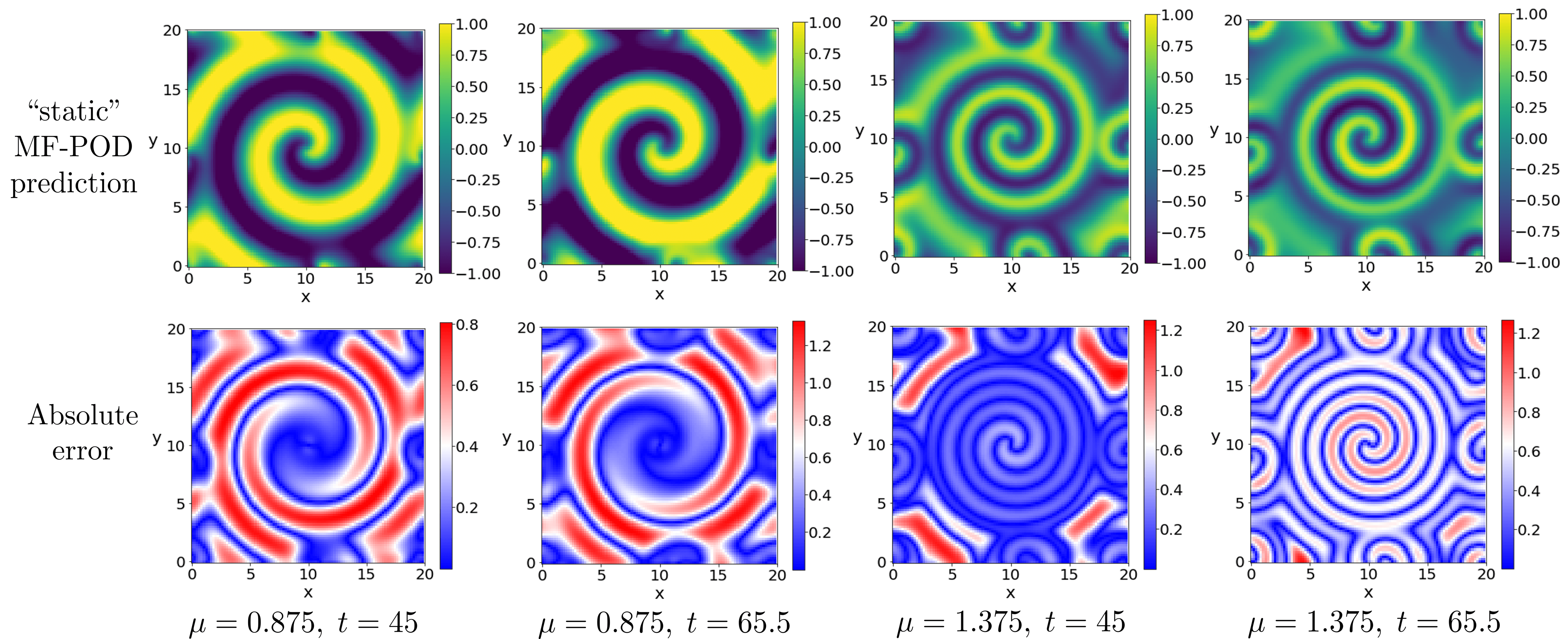}
    \caption{Predicted solution fields in example (I) by  MF-POD with ``static" feed-forward neural networks, instead of LSTM networks. The snapshots refer to two extrapolated time instances $t \in \{45, 65.5\}$ for two testing parameter values $\mu \in \{0.875, 1.375\}$. Absolute error shows the discrepancy between the ``static"  MF solution and the HF reference.}
    \label{fig:static_comparison}
\end{figure}

Although the presented results have shown the proposed method's capability in mapping LF solutions to HF ones, the need to create this map via an LSTM neural network, instead of alternative regression techniques, has not been made clear yet. To show more evidence in this regard, we report further results obtained by replacing the LSTM neural network with a ``static" feed-forward neural network (i.e., without $t$ in the inputs) in the MF regression step (Step 4 offline). In this scenario, the LF POD coefficients are mapped  to their HF counterparts instant by instant, instead of being processed as time series.
In Fig.~\ref{fig:static_comparison}, we report such computed spatial reconstructions and their absolute errors at the same time-parameter test locations considered before. We note a clear worsening in reconstruction accuracy and a significant increase in approximation errors, compared with the predictions by MF-POD with LSTM networks reported in Fig.~\ref{fig:physical_rec_RD}.
This highlights that the MF regression with LSTM networks allows a better detection of temporal patterns in time series and nonlinear correlations between datasets at different fidelity levels, guaranteeing an improved predictive performance. We refer to \cite{conti2023multi} for further discussions on these aspects.

\section{Numerical example II: Advection-diffusion in shallow water}
\label{sect: results2}

In this example, we consider an advection-diffusion problem describing a fluid motion in the shallow water limit \cite{kutz2013data} given by
\begin{subequations}\label{eq: SW_eqs}
\begin{align}
\frac{\partial\omega}{\partial t} + \mu \left( \frac{\partial \psi}{\partial x}\frac{\partial \omega}{\partial y} -  \frac{\partial \psi}{\partial y}\frac{\partial \omega}{\partial x}\right) &= d \nabla^2\omega, \label{eq: SW_eq_1}\\
\nabla^2\psi &= \omega, \label{eq: SW_eq_2}
\end{align}
\end{subequations}
defined over a spatial domain $(x,y)\in [-L,L]^2$ and a time span $t\in [0,T]$. Here $\omega(x,y,t)$ and $\psi(x,y,t)$ represent the vorticity and streamfunction, respectively, $\nabla^2 = \partial_x^2+\partial_y^2$ is the two-dimensional Laplacian,  $d=0.001$ is the diffusion coefficient, and we take $L=10$ and $T=20$. We assume periodic boundary conditions and a stretched Gaussian function as the initial condition of vorticity:
\begin{equation}
    \omega(x,y,0) = \exp\left(-2x^2-\frac{y^2}{20}\right), \qquad (x,y)\in [-L,L]^2\,.
\label{eq: w0}
\end{equation}
We are interested in approximating the time-dependent vorticity field $\omega$ as the parameter $\mu$ varies over $\mathcal{P}=[1,5]$.

The general procedure for solving the system \eqref{eq: SW_eqs} numerically is to \textit{(i)} compute the streamfunction $\psi(x,y,0)$ at $t=0$ by solving the elliptic equation \eqref{eq: SW_eq_2} given the initial condition of vorticity \eqref{eq: w0}, \textit{(ii)} use a time-stepper on \eqref{eq: SW_eq_1} to advance the vorticity $\omega$ by one step $\Delta t$, and \textit{(iii)} repeat \textit{(i)} and \textit{(ii)} starting with the updated $\omega(x,y,t+\Delta t)$ until the final time $T$ is reached. A small time-step $\Delta t$ implies a large number of iterations, and a fine spatial discretization requires solving a large linear system in \textit{(i)}. 
Moreover, this procedure must be repeated for each instance of parameter $\mu$, which makes HF simulations impractical. We use the proposed MF-POD method to relieve these heavy computational burdens. In particular, we consider LF solutions on a coarser spatial grid with larger time-steps, while only evaluating a limited number of HF solution data at several parameter locations for MF-POD training.

\subsection{Multi-fidelity setting}

\begin{figure}[t]
    \centering
    \includegraphics[width=.95\linewidth]{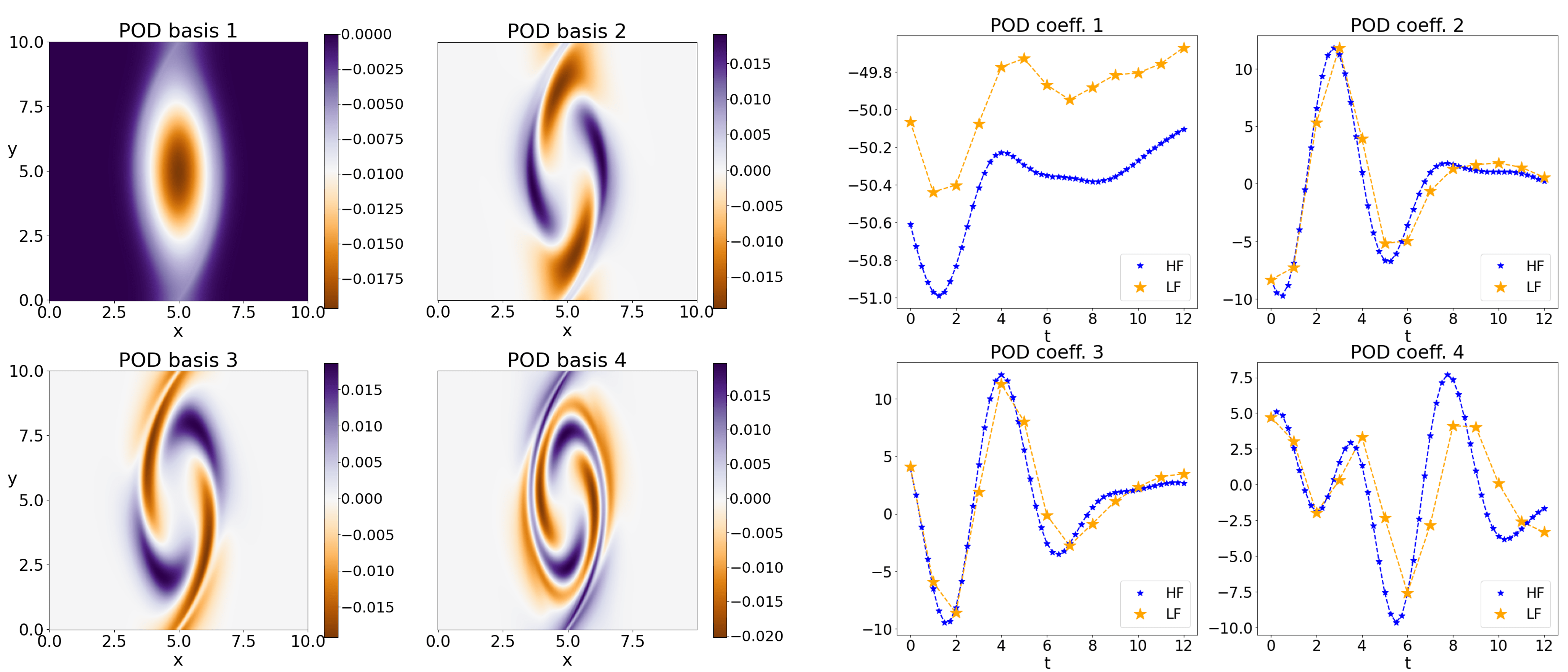}
    \caption{The first four spatial modes identified by the POD (left) 
    and the bi-fidelity data of corresponding expansion coefficients (right) in example (II) 
    ($\mu = 3$).
    Note that the training data only cover a limited time window $[0,T_\text{train}]=[0,12]$ with $T_\text{train}<T = 20$.}
    \label{fig:SW_train}
\end{figure}
As in the previous example, we adopt two fidelity levels that differ in the spatial resolution of discretization, but also consider larger steps of time integration in the LF model than its HF counterpart. 
Specifically, the HF (resp. LF) solution data are generated via the Fourier spectral method on an equispaced spatial grid of $n_\texttt{HF} = 200$ (resp. $n_\texttt{LF} = 50$) nodal points along each direction with time-step size $\Delta t_\texttt{HF}= 0.25$ (resp. $\Delta t_\texttt{LF} = 1.00$). The bi-fidelity training data cover a limited time window $[0,T_\text{train}] = [0,12]$ with $T_\text{train}<T = 20$, only at a small number ($N_\mu = 5$) of parameters locations equispaced over $\mathcal{P}$. 
Thereafter, the training data are fed to the offline algorithm of MF-POD. In this example, we retain the first $N_\texttt{POD} = 17$ POD modes and consider linear interpolation both in time and space to lift LF data to the HF dimensionality for the POD projection. The first four POD modes and the corresponding time-dependent expansion coefficients at $\mu=3$ are depicted in Fig. \ref{fig:SW_train}. 
The predictive performance of MF-POD is tested for $N_\mu^\text{test} = 4$ unseen parameters values $\mu \in \{1.5, 2.5, 3.5, 4.5\}\subset \mathcal{P}$.

\subsection{Results}
The MF-POD method allows us to create an efficient, reliable, low-dimensional surrogate model for the advection-diffusion in shallow water.
As presented in Table~\ref{tab: tab_ex2}, parametric solutions can be evaluated with MF-POD at a computational cost comparable to that of the LF, yet with a significant improvement in accuracy as highlighted by the substantial reduction in predictive error.

This example is challenging for data-driven surrogate modeling, especially in terms of the approximation of POD coefficients describing a parametric coherent structure that propagates over time (i.e., the wave-type phenomena). If only the time-parameter inputs are accounted for, the regression may very likely suffer from difficulties with limited data, leading to poor generalization performance, and extrapolation beyond the training time window would thus be barely possible. The proposed MF-POD method mitigates such technical risks by incorporating physically meaningful LF solutions, which can capture unseen dynamical characteristics and inform the  solution approximation via the LF-to-HF mapping on POD coefficients. 

\begin{figure}[t]
    \centering
    \includegraphics[width=.95\linewidth]{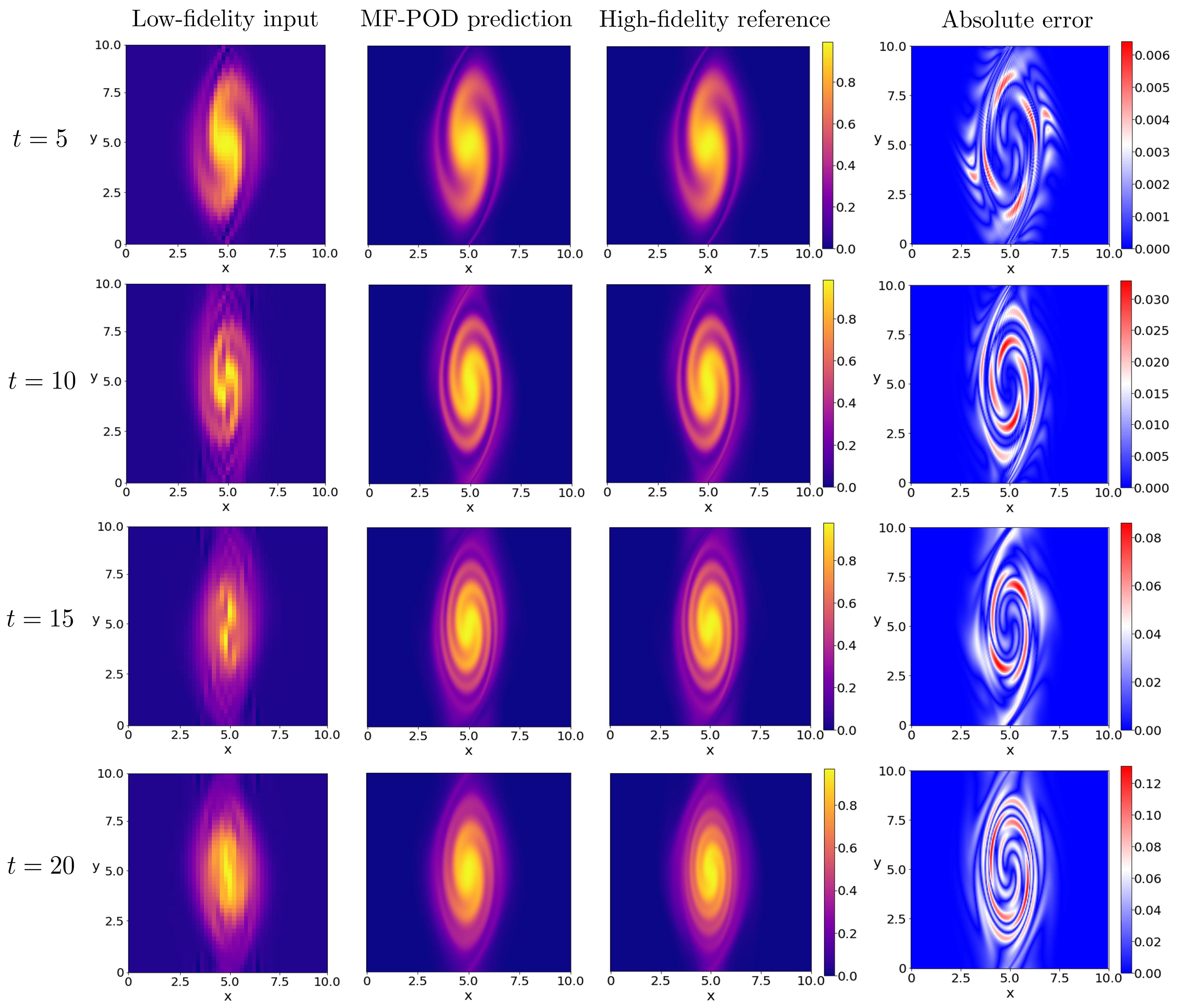}
    \caption{Comparison of solution fields in example (II) among the approximation by MF-POD, the corresponding LF input, and the HF ground truth (used as reference). The snapshots refer to two training time instances $t \in \{5,10\}$  and two extrapolated time instances $t \in \{15, 20\}$ (being $T_\text{train} = 12$ the end of HF training coverage)  for the testing parameter $\mu = 3.5$ that is unseen during the training. Absolute error shows the discrepancy between the MF solution and the HF reference.}
    \label{fig:SW_test}
\end{figure}

To illustrate the effectiveness of MF-POD in predicting forward in time and for new parameter instances simultaneously, we depict in Fig. \ref{fig:SW_test} the evolution of vorticity $\omega$ at an unseen testing parameter value $\mu= 3.5$. The physical behavior is accurately predicted by the MF-POD approach, even with very low resolution on the LF level. As time evolves, the wave propagation presents finer spatial patterns, and the temporal extrapolation becomes more complex, especially considering that no HF information is available after $t=T_\text{train}=12$, and the LF input does not have sufficient resolution to perfectly detect fine patterns. Nevertheless, the proposed method allows to accurately extrapolate (see, e.g., the prediction at $t=15$ in Fig.~\ref{fig:SW_test}) up to time $T=20$, showing the power of MF data fusion, though such power is not unlimited as the predictive error in wave propagation is no longer negligible after $T$.

\begin{table}
\caption{Comparison of computational time and accuracy in example (II).}
\centering
\begin{tabular}{c|c|c|c}
\hline
                   & Low-fidelity input & MF-POD predicted & High-fidelity reference\\ \hline\hline
Computational time &    3.23s (1.03\%)          &      4.12s (1.32\%)          &     311s (100\%)       \\ \hline
Relative error           & 12.8\%    &      3.35\%         &      -        \\ \hline
\end{tabular}
\label{tab: tab_ex2}
\end{table}

\section{Numerical example III: Navier-Stokes equations}
\label{sect: results3}

\begin{figure}[t]
    \centering
    \includegraphics[width=0.8\textwidth]{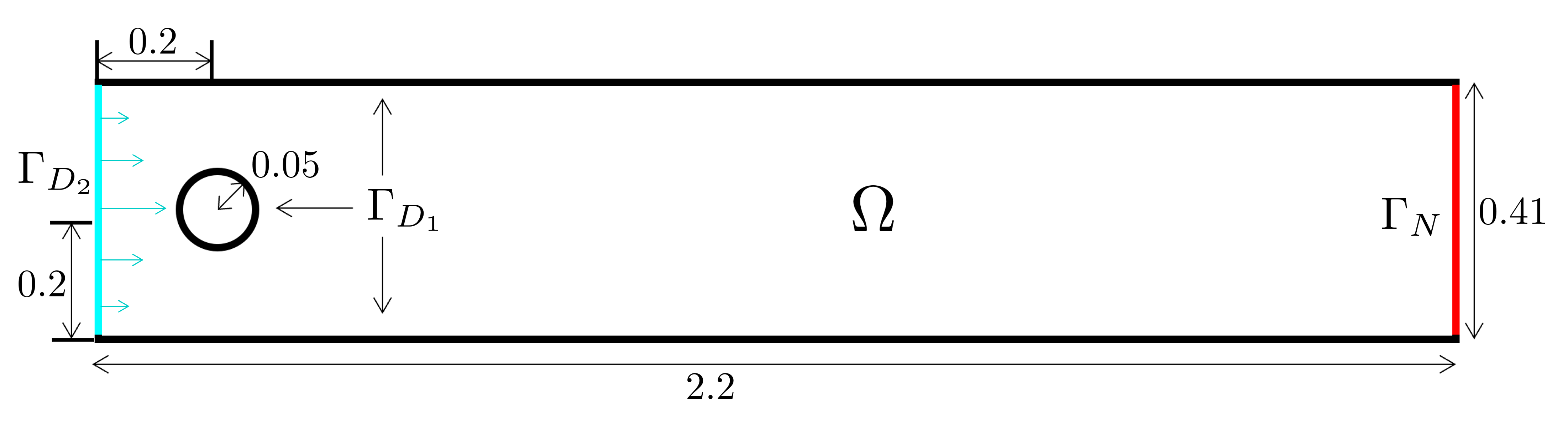}
    \caption{Geometry for a 2D channel flow around a cylinder. All lengths are measured in meters.}
    \label{fig:NS_geometry}
\end{figure}

For the last example, we consider a two-dimensional fluid flow around a cylinder --- a benchmark problem in computational fluid dynamics. Our goal is to efficiently approximate the velocity and pressure fields of a viscous, incompressible Newtonian fluid flow as its Reynolds number varies. The problem is governed by the following Navier-Stokes equations
\begin{equation}
    \begin{aligned}
           \rho \frac{\partial \textbf{v}}{\partial t} - \rho \textbf{v}\cdot \nabla\textbf{v} - \nabla \cdot \bm{\sigma}(\textbf{v},p) &=  \textbf{0}, \qquad &&(\bm{x},t)\in \Omega \times (0,T) \, ,  \\
    \nabla \cdot \textbf{v} &= 0, \qquad &&(\bm{x},t)\in \Omega \times (0,T) \, ,
    \end{aligned}
    \label{eq: NS}
\end{equation}
where $\textbf{v}(\bm{x},t)$ and $p(\bm{x},t)$ represent the velocity and pressure field, respectively, and $\rho = 1.0 \text{ kg/m}^3$ is the fluid density, $\bm{\sigma}(\vb{v},p)=-p\vb{I}+2\nu\bm{\epsilon}(\vb{v})$ is the stress tensor with $ \bm{\epsilon}(\vb{v})$ denoting the strain tensor. The kinematic viscosity is defined as $\nu = 1 / Re$ \cite{fresca2021real}, in which the Reynolds number, $Re$, is the system parameter of interest. As initial conditions we consider the fluid at rest
\begin{equation*}
     \textbf{v}(\bm{x}, 0) = \textbf{0}, \qquad \bm{x} \in \Omega \, ,
\end{equation*}
and we provide the following boundary conditions for the domain $\Omega = (0, 2.2) \times (0, 0.41) \text{\textbackslash} B_r(0.2,0.2)$ ($r = 0.05$), representing a 2D channel with a cylindrical obstacle (see Fig.~\ref{fig:NS_geometry}):
\begin{equation*}
  \begin{aligned}
     \textbf{v} &= \textbf{0}, \qquad &&(\bm{x},t)\in \Gamma_{\text{D}_1} \times (0,T) \, , \\
    \textbf{v} &= \textbf{h}, \qquad &&(\bm{x},t)\in \Gamma_{\text{D}_2} \times (0,T)  \, , \\ 
    \bm{\sigma}(\textbf{v},p)\textbf{n} &= \textbf{0}, \qquad &&(\bm{x},t)\;\in \Gamma_\text{N} \times (0,T)  \, ,
  \end{aligned}
\end{equation*}
which include a no-slip condition on $\Gamma_{\text{D}_1}$, a parabolic inflow 
\[ \textbf{h}(\bm{x},t) = \left(\frac{4U(t)x_2(0.41-x_2)}{0.41^2},0\right)\,,  \qquad 
U(t)=\left\{
  \begin{array}{@{}ll@{}}
    0.75(1-\cos{(\pi t)}), & t <1 \\
    1.5, & t \geq 1
  \end{array}\right.
  \]
on the inlet $\Gamma_{\text{D}_2}$, and an open boundary condition at the outlet $\Gamma_\text{N}$.

In the present study, we consider $\mu = Re \in \mathcal{P} = [30,100]$. When $Re <49$, the flow presents a \textit{laminar} behavior; for larger values of Reynolds number, the flow transitions to an \textit{unsteady} state and a pair of vortices form in the wake of the cylinder, oscillating periodically between the top and bottom sides \cite{RAJANI20091228, zdravkovich1997flow}.
As $Re$ varies across laminar and unsteady ranges, we use the MF-POD method to construct efficient MF surrogate models for the velocity and pressure fields up to $T = 18~\text{s}$, at which time the fluid is fully developed and presents a periodic behavior. 

\subsection{Multi-fidelity setting}

Solution data are generated through a finite element approximation of \eqref{eq: NS} with the backward differentiation formula provided by the MATLAB library \texttt{redbKIT}  \cite{negri2016redbkit}.  We consider a training set computed over a short time window $T_\text{train} = 12~\text{s} <T = 18 ~\text{s}$ at $N_\mu = 15$ Reynolds numbers equispaced over $\mathcal{P}$.
In this case, the difference between the fidelity levels incorporates all the characteristics considered so far in the previous examples, namely the choice of spatial mesh size and time step, as well as whether the physical characteristic $Re$ is corrupted.

For the HF data, the total number of spatial degrees of freedom is $N_\texttt{HF}^\text{dof}=73131$, obtained with quadratic finite elements for the velocity field and linear finite elements for the pressure field over a mesh with 16478 triangular elements and 8239 vertices, while the temporal discretization is with step size $\Delta T_\text{train} = 5 ~\text{ms}$. Instead, LF data are computed with a larger time step $\Delta T_\texttt{LF} = 50~\text{ms}$ over a coarser mesh consisting of 7789 triangular elements and 3899 nodes, thus resulting in snapshots with $N_\texttt{LF}^\text{dof}=34439$. The LF solutions can be computed with  significantly reduced time in comparison with the HF ones. Moreover, we consider a corruption factor of $\alpha = 0.95$ that multiplies $Re$ in the generation of LF data.
For example, when $Re = 60$ for the HF level, the LF solution is evaluated at $\tilde{Re} = \alpha Re  = 57$.

Once solution data are prepared, POD reduction is applied to the HF snapshots and the first  $N_\texttt{POD} = 32$ modes are retained. 
LF data are lifted by linear interpolation over the spatial and temporal domains, and the POD coefficients, obtained by projecting the HF and lifted LF data onto the reduced basis, are then passed to the MF LSTM network for training.

\subsection{Results}
The MF-POD method is tested for approximating the fluid velocity and pressure fields at unseen parameter values $Re \in \{37, 48, 63, 78, 92\}$, while simultaneously extrapolating into the time window $[T_\text{train},T] = [12~\text{s}, 18~\text{s}]$ over which no training data are provided. For a given new value of $Re$, the LF solution is evolved up to the final time of interest $T = 18~\text{s}$, then the LF POD coefficients are computed and mapped through the LSTM network to their HF counterparts, from which the whole velocity and pressure fields are reconstructed.

\begin{figure}[t]
    \centering
    \begin{minipage}[b]{1.\linewidth}
        \centering
        \includegraphics[width=\linewidth]{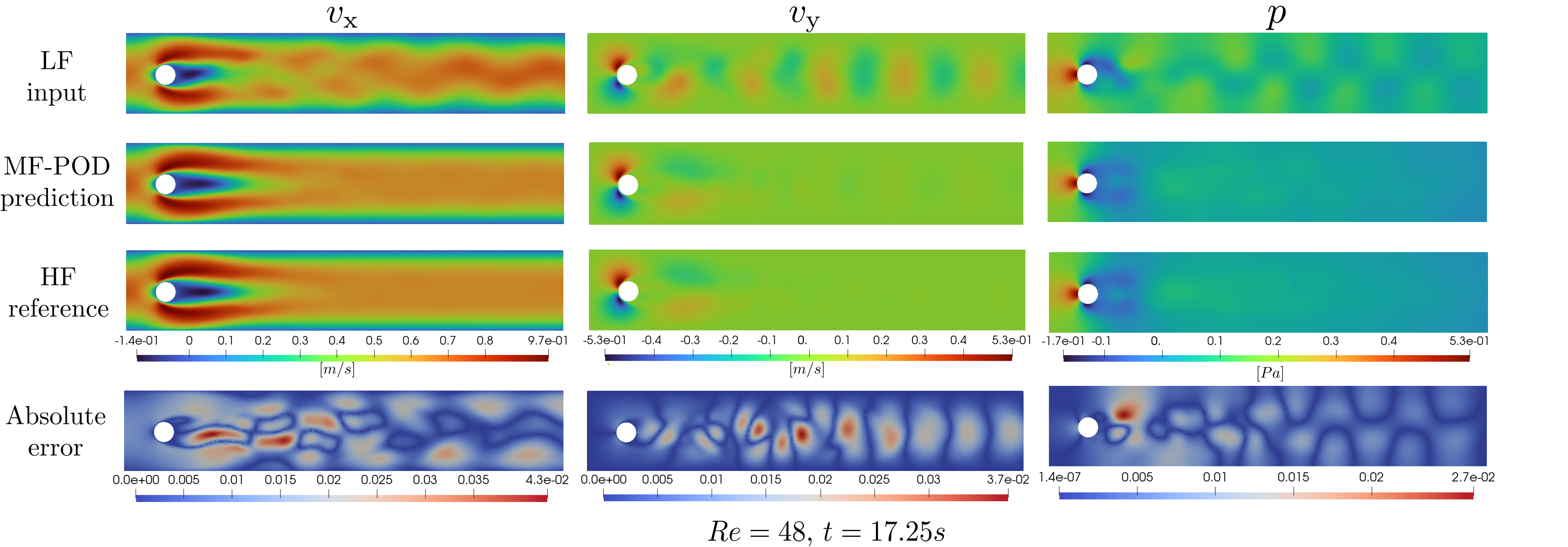}
    \end{minipage}
    \hfill
    \vspace{2pt}
    \begin{minipage}[b]{1.\linewidth}
        \centering
        \includegraphics[width=\linewidth]{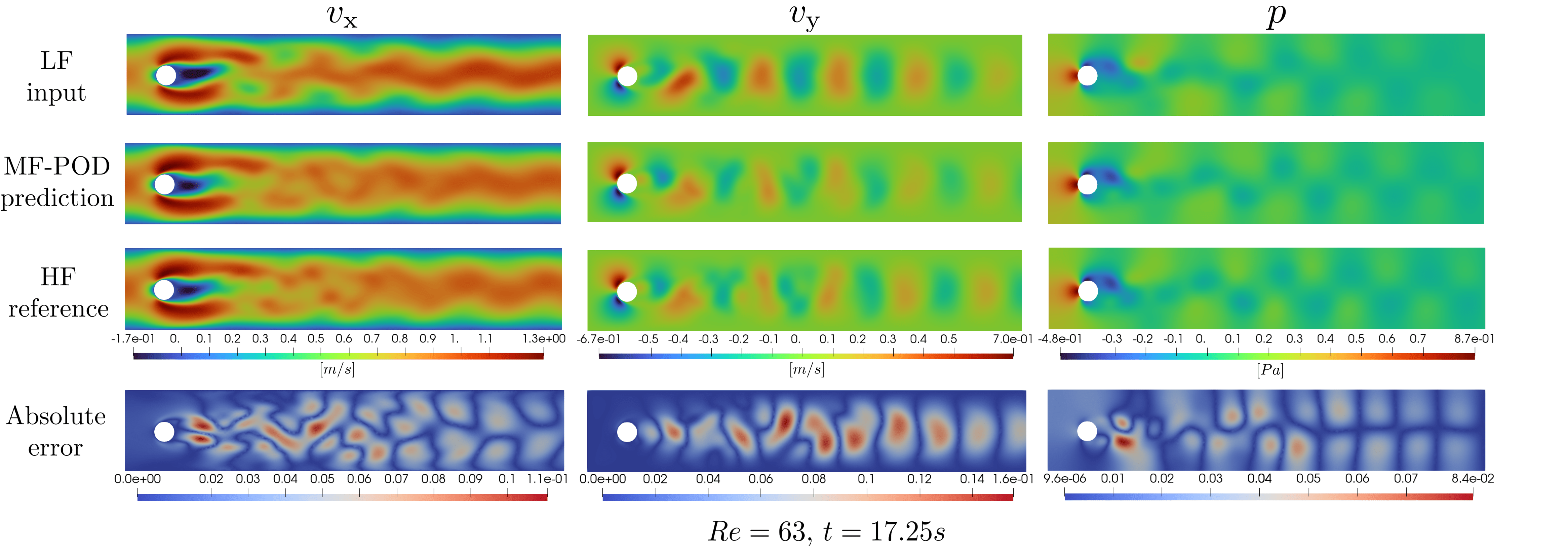}
        \caption{Comparison of solution fields in example (III) among the approximation by MF-POD, the corresponding LF input, and the HF ground truth (used as reference). The snapshots refer to the extrapolated time instance $t = 17.25~\text{s}$ (being $T_\text{train} = 12~\text{s}$ the end of HF training coverage) for two testing values of Reynolds number $Re = 48$ (above) and $64$ and (below), both of which are unseen during the training. Absolute error shows the discrepancy between the MF solution and the HF reference.}
        \label{fig: physical_rec_NS}
    \end{minipage}
\end{figure}

\begin{figure}[t]
    \centering
    \includegraphics[width=1.\linewidth]{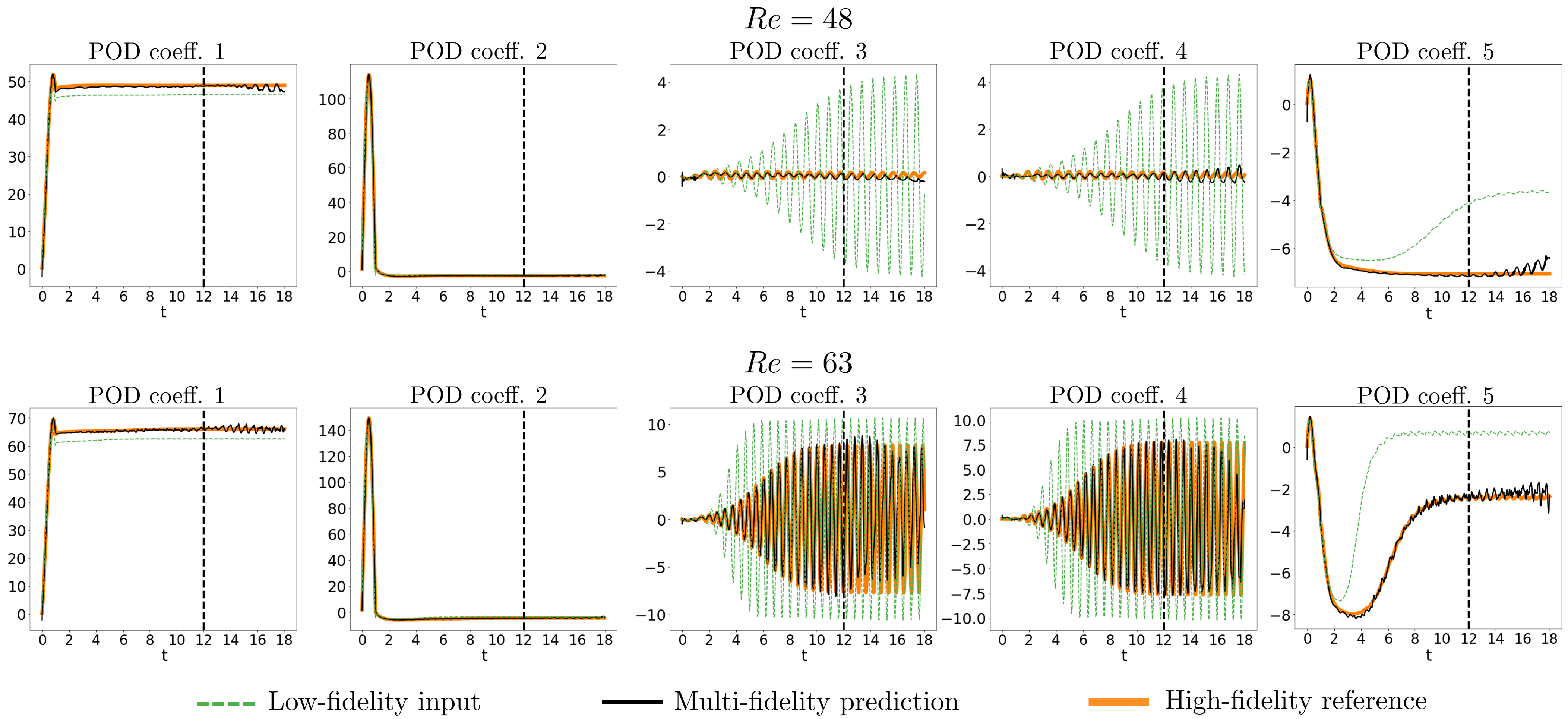}
    \caption{Comparison of the time evolution of POD coefficients in example (III) among MF solution, LF input, and HF reference at testing parameter instances $Re = 48$ (above) and  $63$ (below). The dashed line at $T_\text{train} = 12~\text{s}$ indicates the end time of HF data coverage in training, i.e., no HF information is available over $t\in[T_\text{train},T] = [12~\text{s},18~\text{s}]$.}
    \label{fig: temporal_coeff_NS}
\end{figure}

In Table \ref{tab: tab_ex3}, we report the computational costs and  predictive errors evaluated over the test set. Once again, we observe that the MF-POD method achieves advantages over the HF and LF models, by drastically reducing computational time while preserving a good accuracy, respectively.
Moreover, in Fig. \ref{fig: physical_rec_NS}, we illustrate the MF predictive solutions for two testing parameter values $Re \in \{48,63\}$ in comparison with their LF inputs and HF references. 
These two $Re$ values correspond to two different regimes of the fluid flow --- laminar and unsteady, respectively. In both cases, the proposed model is able to capture correct fluid behaviors with a good accuracy. 
For Reynolds numbers close to the bifurcation ($Re = 49$) between laminar and unsteady regimes, a coarse approximation may lead to significant errors in simulating the dynamical behaviors of the fluid. For example, when $Re = 48$, we observe that the LF solution exhibits the onset of unsteady oscillatory phenomena, which are, instead, absent in the reference HF solution, which exhibits the expected laminar behavior. This is even more evident by observing the temporal evolution of POD coefficients as shown in Fig. \ref{fig: temporal_coeff_NS}, where we notice that the LF solution features a non-negligible oscillatory contribution from the third and fourth POD coefficients; these latter are instead almost vanishing 
for the reference HF solution. We note that MF-POD is able to accurately recover the correct HF characteristics.  

\begin{table}[htb!]
\caption{Comparison of computational time and accuracy in example (III).}

\centering
\begin{tabular}{c|c|c|c}
\hline
                   & Low-fidelity input & MF-POD predicted & High-fidelity reference \\ \hline\hline
Computational time &    108s (4.05\%)          &     111s (4.17\%)          &     2664s (100\%)       \\ \hline
Relative error           &  18.5\%     &     3.48\%          &     -         \\ \hline
\end{tabular}
\label{tab: tab_ex3}
\end{table}

\section{Conclusions}
\label{sect: conclusions}

\begin{figure}[b]
    \centering
    \includegraphics[width=1.\linewidth]{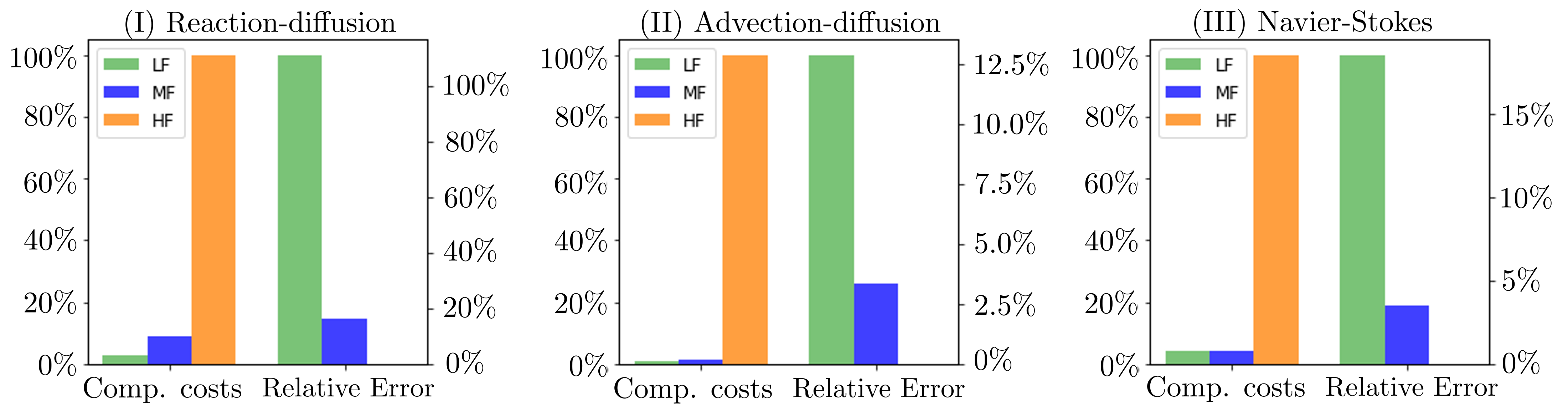}
    \caption{Comparison of the LF, HF, and the proposed MF approaches in terms of computational times and accuracy. For each considered example, we provide a plot with the visualizations of the values collected in Tables \ref{tab: tab_ex1}-\ref{tab: tab_ex2}-\ref{tab: tab_ex3}. Values of the computational costs are indicated on the left vertical axis as percentages with respect to the computational costs of running HF solvers, while relative prediction errors with respect to the HF reference solutions are indicated on the vertical axis on the right. We observe how the proposed MF method allows for a significant improvement in predictive accuracy with respect to the LF approach, while maintaining extremely limited online computational costs, in particular as compared to the HF ones.}
    \label{fig: barplot}
\end{figure}

We have developed a new reduced-order surrogate modeling method that relies on solution data from different fidelity levels to reduce computational costs while preserving predictive accuracy. 
Like traditional model reduction, the MF-POD architecture employs POD to extract a low-dimensional basis that approximates the solution manifold from a limited amount of HF data.  An  MF LSTM model is subsequently trained to infer the temporal evolution of the HF solution on the POD manifold from its LF counterpart. 
Once trained \textit{offline}, 
the proposed model can be deployed \textit{online} to generate new solutions that approach HF accuracy at the computational cost of LF evaluations. The advantages of the MF POD technique has been compared to both HF and LF methods directly through a diverse number of example PDEs with the results summarized in Fig.~\ref{fig: barplot}.

By incorporating physically meaningful LF data, MF-POD addresses the potential lack of physical consistency in purely data-driven methods, thereby ensuring interpretability and reliability in parametric generalization and temporal forecasting while still benefiting from its non-intrusive nature.
A limitation of the proposed framework is the assumption of a strict hierarchy among fidelity levels, which implies that LF solutions are required at parameter configurations in which we are interested on the HF level. A further limitation is that this method has not fully benefited from the large availability of LF data. In fact, the MF LSTM networks are trained on the same amount of LF and HF data to learn the LF-to-HF mapping, sampled at the same parameter locations (see Fig.~\ref{fig:train}). Moreover, the low-dimensional basis is constructed solely with the HF data. Therefore, potential LF information could be additionally exploited to improve both the construction of reduced basis and the MF regression for the expansion coefficients.

However, the non-intrusiveness of the proposed method guarantees remarkable flexibility
in numerical implementations, and enables seamless adaptation, extension, and enhancement of its individual components. For instance, the use of other techniques for reduced basis construction, which may be better tailored to suit specific applications of interest, can replace the POD reduction. 
Moreover, alternative recurrent neural network  or other emerging architectures (e.g., transformers \cite{vaswani2017attention}) can be employed instead of the LSTM layers in the MF regression task. Such refinements will further empower the proposed method to evolve and excel in diverse physical simulations in computational science and engineering.

In particular, as a direction for future development, the presented framework can be naturally extended to real-life applications, in which experimental measurement data can be fused with synthetic (LF) simulation data to create  surrogate models that produce accurate real-time predictions beyond the experimental data coverage.

\section*{Acknowledgment}
PC is supported under the JRC STEAM STM-Politecnico di Milano agreement. 
MG receives support from Sectorplan Bèta (the Netherlands) under the focus area \emph{Mathematics of Computational Science}. 
The present research has been partially supported by FAIR (Future Artificial Intelligence Research) project, funded by the NextGenerationEU program within the PNRR-PE-AI scheme (M4C2, Investment 1.3, Line on Artificial Intelligence) and by MUR, grant Dipartimento di Eccellenza 2023-2027. PC, SLB, and JNK acknowledge generous funding support from the National Science Foundation AI Institute in Dynamic Systems
(grant number 2112085). 
 
\bibliographystyle{abbrv}
\bibliography{references.bib}

\begin{thebibliography}{10}

\bibitem{ahmed2021multifidelity}
S.~E. Ahmed, O.~San, K.~Kara, R.~Younis, and A.~Rasheed.
\newblock Multifidelity computing for coupling full and reduced order models.
\newblock {\em Plos one}, 16(2):e0246092, 2021.

\bibitem{alvarez2012kernels}
M.~A. {\'A}lvarez, L.~Rosasco, and N.~D. Lawrence.
\newblock Kernels for vector-valued functions: A review.
\newblock {\em Foundations and Trends{\textregistered} in Machine Learning},
  4(3):195--266, 2012.

\bibitem{Antoulas2005}
A.~C. Antoulas.
\newblock A survey of model reduction methods for large-scale systems.
\newblock {\em Contemporary Mathematics}, 280:193--219, 2005.

\bibitem{bakarji2022discovering}
J.~Bakarji, K.~Champion, J.~N. Kutz, and S.~L. Brunton.
\newblock Discovering governing equations from partial measurements with deep
  delay autoencoders.
\newblock {\em arXiv preprint arXiv:2201.05136}, 2022.

\bibitem{benner2015survey}
P.~Benner, S.~Gugercin, and K.~Willcox.
\newblock A survey of projection-based model reduction methods for parametric
  dynamical systems.
\newblock {\em SIAM review}, 57(4):483--531, 2015.

\bibitem{bergstra2011algorithms}
J.~Bergstra, R.~Bardenet, Y.~Bengio, and B.~K{\'e}gl.
\newblock Algorithms for hyper-parameter optimization.
\newblock {\em Advances in Neural Information Processing Systems}, 24, 2011.

\bibitem{bergstra2022hyperopt}
J.~Bergstra, D.~Yamins, and D.~Cox.
\newblock Hyperopt: Distributed asynchronous hyper-parameter optimization.
\newblock {\em Astrophysics Source Code Library}, pages ascl--2205, 2022.

\bibitem{botteghi2022deep}
N.~Botteghi, M.~Guo, and C.~Brune.
\newblock Deep kernel learning of dynamical models from high-dimensional noisy
  data.
\newblock {\em Scientific reports}, 12(1):21530, 2022.

\bibitem{brunton2022data}
S.~L. Brunton and J.~N. Kutz.
\newblock {\em Data-driven science and engineering: Machine learning, dynamical
  systems, and control}.
\newblock Cambridge University Press, 2022.

\bibitem{brunton2016discovering}
S.~L. Brunton, J.~L. Proctor, and J.~N. Kutz.
\newblock Discovering governing equations from data by sparse identification of
  nonlinear dynamical systems.
\newblock {\em Proceedings of the national academy of sciences},
  113(15):3932--3937, 2016.

\bibitem{bui2008parametric}
T.~Bui-Thanh, K.~Willcox, and O.~Ghattas.
\newblock Parametric reduced-order models for probabilistic analysis of
  unsteady aerodynamic applications.
\newblock {\em AIAA journal}, 46(10):2520--2529, 2008.

\bibitem{MDEIM}
K.~T. Carlberg, R.~Tuminaro, and P.~Boggs.
\newblock Preserving lagrangian structure in nonlinear model reduction with
  application to structural dynamics.
\newblock {\em SIAM J. Sci. Comput}, 37(2):B153--B184, 2015.

\bibitem{champion2019sindy}
K.~Champion, B.~Lusch, J.~N. Kutz, and S.~L. Brunton.
\newblock Data-driven discovery of coordinates and governing equations.
\newblock {\em Proceedings of the National Academy of Sciences},
  116(45):22445--22451, 2019.

\bibitem{MFPOD_repo}
P.~Conti.
\newblock {MultiFidelity\_POD}.
\newblock \url{https://github.com/ContiPaolo/MultiFidelity_POD}, 2023.

\bibitem{conti2023reduced}
P.~Conti, G.~Gobat, S.~Fresca, A.~Manzoni, and A.~Frangi.
\newblock Reduced order modeling of parametrized systems through autoencoders
  and {S}{I}{N}{D}y approach: continuation of periodic solutions.
\newblock {\em Computer Methods in Applied Mechanics and Engineering},
  411:116072, 2023.

\bibitem{conti2023multi}
P.~Conti, M.~Guo, A.~Manzoni, and J.~S. Hesthaven.
\newblock Multi-fidelity surrogate modeling using long short-term memory
  networks.
\newblock {\em Computer methods in applied mechanics and engineering},
  404:115811, 2023.

\bibitem{cui2015data}
T.~Cui, Y.~Marzouk, and K.~Willcox.
\newblock Data-driven model reduction for the bayesian solution of inverse
  problems.
\newblock {\em International Journal for Numerical Methods in Engineering},
  102(5):966--990, 2015.

\bibitem{frangos2010surrogate}
M.~Frangos, Y.~Marzouk, K.~Willcox, and B.~van Bloemen~Waanders.
\newblock Surrogate and reduced-order modeling: a comparison of approaches for
  large-scale statistical inverse problems.
\newblock {\em Large-Scale Inverse Problems and Quantification of Uncertainty},
  pages 123--149, 2010.

\bibitem{fresca2020comprehensive}
S.~Fresca, L.~Dede, and A.~Manzoni.
\newblock A comprehensive deep learning-based approach to reduced order
  modeling of nonlinear time-dependent parametrized pdes.
\newblock {\em Journal of Scientific Computing}, 87(2):1--36, 2021.

\bibitem{fresca2021real}
S.~Fresca and A.~Manzoni.
\newblock Real-time simulation of parameter-dependent fluid flows through deep
  learning-based reduced order models.
\newblock {\em Fluids}, 6(7):259, 2021.

\bibitem{fresca2021poddlrom}
S.~Fresca and A.~Manzoni.
\newblock {P}{O}{D}-{D}{L}-{R}{O}{M}: enhancing deep learning-based reduced
  order models for nonlinear parametrized pdes by proper orthogonal
  decomposition.
\newblock {\em Computer Methods in Applied Mechanics and Engineering},
  388:114181, 2022.

\bibitem{geneva2020multi}
N.~Geneva and N.~Zabaras.
\newblock Multi-fidelity generative deep learning turbulent flows.
\newblock {\em Foundations of Data Science}, 2(4):391--428, 2020.

\bibitem{gers2000learning}
F.~A. Gers, J.~Schmidhuber, and F.~Cummins.
\newblock Learning to forget: Continual prediction with lstm.
\newblock {\em Neural Computation}, 12(10):2451--2471, 2000.

\bibitem{ghattas2021learning}
O.~Ghattas and K.~Willcox.
\newblock Learning physics-based models from data: perspectives from inverse
  problems and model reduction.
\newblock {\em Acta Numerica}, 30:445--554, 2021.

\bibitem{guo2019data}
M.~Guo and J.~S. Hesthaven.
\newblock Data-driven reduced order modeling for time-dependent problems.
\newblock {\em Computer methods in applied mechanics and engineering},
  345:75--99, 2019.

\bibitem{guo2022multi}
M.~Guo, A.~Manzoni, M.~Amendt, P.~Conti, and J.~S. Hesthaven.
\newblock Multi-fidelity regression using artificial neural networks: efficient
  approximation of parameter-dependent output quantities.
\newblock {\em Computer methods in Applied Mechanics and Engineering},
  389:114378, 2022.

\bibitem{guo2022bayesian}
M.~Guo, S.~A. McQuarrie, and K.~E. Willcox.
\newblock Bayesian operator inference for data-driven reduced-order modeling.
\newblock {\em Computer Methods in Applied Mechanics and Engineering},
  402:115336, 2022.

\bibitem{haik2023real}
W.~Haik, Y.~Maday, and L.~Chamoin.
\newblock A real-time variational data assimilation method with data-driven
  model enrichment for time-dependent problems.
\newblock {\em Comput. Methods Appl. Mech. Engrg.}, 405:115868, 2023.

\bibitem{HRS}
J.~S. Hesthaven, G.~Rozza, and B.~Stamm.
\newblock {\em Certified reduced basis methods for parametrized partial
  differential equations}.
\newblock Springer International Publishing, 2016.

\bibitem{hochreiter1997long}
S.~Hochreiter and J.~Schmidhuber.
\newblock Long short-term memory.
\newblock {\em Neural Computation}, 9(8):1735--1780, 1997.

\bibitem{kast2020non}
M.~Kast, M.~Guo, and J.~S. Hesthaven.
\newblock A non-intrusive multifidelity method for the reduced order modeling
  of nonlinear problems.
\newblock {\em Comput. Methods Appl. Mech. Engrg.}, 364:112947, 2020.

\bibitem{kennedy2000predicting}
M.~C. Kennedy and A.~O'Hagan.
\newblock Predicting the output from a complex computer code when fast
  approximations are available.
\newblock {\em Biometrika}, 87(1):1--13, 2000.

\bibitem{kingma2014adam}
D.~P. Kingma and J.~Ba.
\newblock Adam: {A} method for stochastic optimization.
\newblock {\em arXiv preprint arXiv:1412.6980}, 2014.

\bibitem{kutz2013data}
J.~N. Kutz.
\newblock {\em Data-driven modeling \& scientific computation: methods for
  complex systems \& big data}.
\newblock Oxford University Press, 2013.

\bibitem{lee2020model}
K.~Lee and K.~T. Carlberg.
\newblock Model reduction of dynamical systems on nonlinear manifolds using
  deep convolutional autoencoders.
\newblock {\em Journal of Computational Physics}, 404:108973, 2020.

\bibitem{liu2019multi}
D.~Liu and Y.~Wang.
\newblock Multi-fidelity physics-constrained neural network and its application
  in materials modeling.
\newblock {\em Journal of Mechanical Design}, 141(12), 2019.

\bibitem{pateraYanoWD}
Y.~Maday, A.~Patera, J.~Penn, and M.~Yano.
\newblock A parametrized-background data-weak approach to variational data
  assimilation: formulation, analysis, and application to acoustics.
\newblock {\em Int. J. Numer. Methods Engrg.}, 102:933--965, 2015.

\bibitem{manzoni2012shape}
A.~Manzoni, A.~Quarteroni, and G.~Rozza.
\newblock Shape optimization for viscous flows by reduced basis methods and
  free-form deformation.
\newblock {\em International Journal for Numerical Methods in Fluids},
  70(5):646--670, 2012.

\bibitem{mars2022bayesian}
L.~Mars~Gao and J.~N. Kutz.
\newblock Bayesian autoencoders for data-driven discovery of coordinates,
  governing equations and fundamental constants.
\newblock {\em arXiv e-prints}, pages arXiv--2211, 2022.

\bibitem{maulik2021reduced}
R.~Maulik, B.~Lusch, and P.~Balaprakash.
\newblock Reduced-order modeling of advection-dominated systems with recurrent
  neural networks and convolutional autoencoders.
\newblock {\em Physics of Fluids}, 33(3):037106, 2021.

\bibitem{meng2020multi}
X.~Meng, H.~Babaee, and G.~E. Karniadakis.
\newblock Multi-fidelity {B}ayesian neural networks: Algorithms and
  applications.
\newblock {\em Journal of Computational Physics}, 438:110361, 2021.

\bibitem{meng2020composite}
X.~Meng and G.~E. Karniadakis.
\newblock A composite neural network that learns from multi-fidelity data:
  Application to function approximation and inverse pde problems.
\newblock {\em Journal of Computational Physics}, 401:109020, 2020.

\bibitem{Motamed}
M.~Motamed.
\newblock A multi-fidelity neural network surrogate sampling method for
  uncertainty quantification.
\newblock {\em International Journal for Uncertainty Quantification}, 10(4),
  2020.

\bibitem{nakamura2021convolutional}
T.~Nakamura, K.~Fukami, K.~Hasegawa, Y.~Nabae, and K.~Fukagata.
\newblock Convolutional neural network and long short-term memory based reduced
  order surrogate for minimal turbulent channel flow.
\newblock {\em Physics of Fluids}, 33(2):025116, 2021.

\bibitem{negri2016redbkit}
F.~Negri.
\newblock redb{KIT} {V}ersion 2.2.
\newblock \url{http://redbkit.github.io/redbKIT/}, 2016.

\bibitem{negri2013reduced}
F.~Negri, G.~Rozza, A.~Manzoni, and A.~Quarteroni.
\newblock Reduced basis method for parametrized elliptic optimal control
  problems.
\newblock {\em SIAM Journal on Scientific Computing}, 35(5):A2316--A2340, 2013.

\bibitem{Noack2011book}
B.~R. Noack, M.~Morzynski, and G.~Tadmor.
\newblock {\em Reduced-order modelling for flow control}, volume 528.
\newblock Springer Science \&amp; Business Media, 2011.

\bibitem{noack2011galerkin}
B.~R. Noack, M.~Schlegel, M.~Morzynski, and G.~Tadmor.
\newblock {\em Galerkin method for nonlinear dynamics}.
\newblock Springer, DMD 2011.

\bibitem{olah2015understanding}
C.~Olah.
\newblock Understanding lstm networks.
\newblock \url{https://colah.github.io/posts/2015-08-Understanding-LSTMs/},
  2015.
\newblock Accessed: April 21, 2023.

\bibitem{otto2019linearly}
S.~E. Otto and C.~W. Rowley.
\newblock Linearly recurrent autoencoder networks for learning dynamics.
\newblock {\em SIAM Journal on Applied Dynamical Systems}, 18(1):558--593,
  2019.

\bibitem{peherstorfer2016data}
B.~Peherstorfer and K.~Willcox.
\newblock Data-driven operator inference for nonintrusive projection-based
  model reduction.
\newblock {\em Computer Methods in Applied Mechanics and Engineering},
  306:196--215, 2016.

\bibitem{peherstorfer2018survey}
B.~Peherstorfer, K.~Willcox, and M.~Gunzburger.
\newblock Survey of multifidelity methods in uncertainty propagation,
  inference, and optimization.
\newblock {\em SIAM Review}, 60(3):550--591, 2018.

\bibitem{pepper2021local}
N.~Pepper, A.~Gaymann, S.~Sharma, and F.~Montomoli.
\newblock Local bi-fidelity field approximation with knowledge based neural
  networks for computational fluid dynamics.
\newblock {\em Scientific Reports}, 11(1):1--11, 2021.

\bibitem{perron2020development}
C.~Perron, D.~Rajaram, and D.~Mavris.
\newblock Development of a multi-fidelity reduced-order model based on manifold
  alignment.
\newblock In {\em AIAA Aviation 2020 Forum}, page 3124, 2020.

\bibitem{qian2020lift}
E.~Qian, B.~Kramer, B.~Peherstorfer, and K.~Willcox.
\newblock Lift \& learn: Physics-informed machine learning for large-scale
  nonlinear dynamical systems.
\newblock {\em Physica D: Nonlinear Phenomena}, 406:132401, 2020.

\bibitem{QMN}
A.~Quarteroni, A.~Manzoni, and F.~Negri.
\newblock {\em Reduced Basis Methods for Partial Differential Equations. An
  Introduction}.
\newblock Springer International Publishing, 2016.

\bibitem{raissi2017inferring}
M.~Raissi, P.~Perdikaris, and G.~E. Karniadakis.
\newblock Inferring solutions of differential equations using noisy
  multi-fidelity data.
\newblock {\em J. Comput. Phys.}, 335:736--746, 2017.

\bibitem{RAJANI20091228}
B.~Rajani, A.~Kandasamy, and S.~Majumdar.
\newblock Numerical simulation of laminar flow past a circular cylinder.
\newblock {\em Appl. Math. Mod.}, 33(3):1228 -- 1247, 2009.

\bibitem{rubio2021real}
P.-B. Rubio, L.~Chamoin, and F.~Louf.
\newblock Real-time data assimilation and control on mechanical systems under
  uncertainties.
\newblock {\em Adv. Model. Simul. Eng. Sci}, 8(1):1--25, 2021.

\bibitem{schaeffer2017learning}
H.~Schaeffer.
\newblock Learning partial differential equations via data discovery and sparse
  optimization.
\newblock {\em Proceedings of the Royal Society A: Mathematical, Physical and
  Engineering Sciences}, 473(2197):20160446, 2017.

\bibitem{schmid2010dynamic}
P.~J. Schmid.
\newblock Dynamic mode decomposition of numerical and experimental data.
\newblock {\em Journal of fluid mechanics}, 656:5--28, 2010.

\bibitem{sinigaglia2022fast}
C.~Sinigaglia, D.~E. Quadrelli, A.~Manzoni, and F.~Braghin.
\newblock Fast active thermal cloaking through pde-constrained optimization and
  reduced-order modelling.
\newblock {\em Proceedings of the Royal Society A}, 478(2258):20210813, 2022.

\bibitem{sudret2000stochastic}
B.~Sudret and A.~Der~Kiureghian.
\newblock {\em Stochastic finite element methods and reliability: a
  state-of-the-art report}.
\newblock Department of Civil and Environmental Engineering, University of
  California~…, 2000.

\bibitem{torzoni2023multi}
M.~Torzoni, A.~Manzoni, and S.~Mariani.
\newblock A multi-fidelity surrogate model for structural health monitoring
  exploiting model order reduction and artificial neural networks.
\newblock {\em Mechanical Systems and Signal Processing}, 197:110376, 2023.

\bibitem{trefethen2000spectral}
L.~N. Trefethen.
\newblock {\em Spectral methods in MATLAB}.
\newblock SIAM, 2000.

\bibitem{vaswani2017attention}
A.~Vaswani, N.~Shazeer, N.~Parmar, J.~Uszkoreit, L.~Jones, A.~N. Gomez,
  {\L}.~Kaiser, and I.~Polosukhin.
\newblock Attention is all you need.
\newblock {\em Advances in Neural Information Processing Systems}, 30, 2017.

\bibitem{vlachas2022multiscale}
P.~R. Vlachas, G.~Arampatzis, C.~Uhler, and P.~Koumoutsakos.
\newblock Multiscale simulations of complex systems by learning their effective
  dynamics.
\newblock {\em Nature Machine Intelligence}, 4(4):359--366, 2022.

\bibitem{vlachas2018data}
P.~R. Vlachas, W.~Byeon, Z.~Y. Wan, T.~P. Sapsis, and P.~Koumoutsakos.
\newblock Data-driven forecasting of high-dimensional chaotic systems with long
  short-term memory networks.
\newblock {\em Proceedings of the Royal Society A: Mathematical, Physical and
  Engineering Sciences}, 474(2213):20170844, 2018.

\bibitem{wang2009general}
C.~Wang and S.~Mahadevan.
\newblock A general framework for manifold alignment.
\newblock In {\em 2009 AAAI Fall Symposium Series}, 2009.

\bibitem{zdravkovich1997flow}
M.~M. Zdravkovich.
\newblock {\em Flow around circular cylinders: Volume 2: Applications},
  volume~2.
\newblock Oxford university press, 1997.

\bibitem{zhuang2021model}
Q.~Zhuang, J.~M. Lorenzi, H.-J. Bungartz, and D.~Hartmann.
\newblock Model order reduction based on {R}unge--{K}utta neural networks.
\newblock {\em Data-Centric Engineering}, 2:e13, 2021.

\end{thebibliography}

\end{document}